\documentclass[review]{elsarticle}
\usepackage{lineno,hyperref}
\modulolinenumbers[5]
\journal{Journal of \LaTeX\ Templates}

\biboptions{authoryear}

\usepackage{amsmath}
\usepackage{amssymb}
\usepackage{amsfonts}
\usepackage{placeins}
\usepackage{threeparttable}
\usepackage{latexsym}
\usepackage{lipsum}
\usepackage{multicol}
\usepackage{accents}

\usepackage{float}
\usepackage{euscript}
\usepackage{color}
\usepackage{hyperref}
\usepackage{setspace}
\usepackage{fancyhdr}

\usepackage{geometry}
\usepackage{textcomp}

\usepackage{algorithm,algpseudocode}
\usepackage{moresize}
\usepackage{psfrag}
\usepackage{subfigure}
\usepackage{gensymb}
\usepackage{colortbl}
\usepackage{multirow}
\usepackage{epstopdf}
\usepackage{cleveref}
\usepackage{tabto}

\usepackage{url}
\usepackage[table]{xcolor}
\usepackage{comment}

\begin{document}
\begin{frontmatter}
\title{Control and Navigation Framework for a Hybrid Steel Bridge Inspection Robot}

\tnotetext[mytitlenote]{This work is supported by the U.S. National Science Foundation (NSF) under grants NSF-CAREER: 1846513 and NSF-PFI-TT: 1919127, and the U.S. Department of Transportation, Office of the Assistant Secretary for Research and Technology (USDOT/OST-R) under Grant No. 69A3551747126 through INSPIRE University Transportation Center. 
 The views, opinions, findings and conclusions reflected in this publication are solely those of the authors and do not represent the official policy or position of the NSF and USDOT/OST-R. \\
 The source code of the work in ROS is available at the ARA lab's github: \url{https://github.com/aralab-unr/ara_navigation}}

\author{Hoang-Dung Bui and Hung Manh La}
\address{Advanced Robotics and Automation
(ARA) Laboratory, Department of Computer Science and Engineering, University of Nevada, Reno, NV 89557, USA.}
\author[mymainaddress,mysecondaryaddress]{ARA Lab.}
\ead[url]{https://ara.cse.unr.edu}

\author[mysecondaryaddress]{University of Nevada\corref{mycorrespondingauthor}}
\cortext[mycorrespondingauthor]{Corresponding author}
\ead{hla@unr.edu}

\begin{abstract}
Autonomous navigation of steel bridge inspection robots is essential for proper maintenance. Majority of existing robotic solutions for steel bridge inspection require human intervention to assist in the control and navigation. In this paper, a control and navigation framework has been proposed for the steel bridge inspection robot developed by the Advanced Robotics and Automation (ARA) to facilitate autonomous real-time navigation and minimize human intervention.
The ARA robot is designed to work in two modes: mobile and inch-worm. The robot uses mobile mode when moving on a plane surface and inch-worm mode when jumping from one surface to the other. To allow the ARA robot to switch between mobile and inch-worm modes, a switching controller is developed with 3D point cloud data based. The surface detection algorithm is proposed to allow the robot to check the availability of steel surfaces (plane, area and height) to determine the transformation from mobile mode to inch-worm one.
To have the robot to safely navigate and visit all steel members of the bridge, four algorithms are developed to process the data from a depth camera, segment it into clusters, estimate the boundaries, construct a graph representing the structure, generate the shortest inspection path with any starting and ending points, and determine available robot configuration for path planning.
Experiments on steel bridge structures setup highlight the effective performance of the algorithms, and the potential to apply to the ARA robot to run on real bridge structures.
\end{abstract}

\end{frontmatter}
\section{Introduction}\label{sec_intro}
Within the field of health monitoring of bridge structures \cite{6917066,s20164403,s20143954,AHMED2020103393}, the development of novel robotic platforms has received considerable attention in the recent years \cite{6705706,5980131,6942821,6653886,6942821,https://doi.org/10.1002/rob.21725,doi:10.1061/9780784479117.032,https://doi.org/10.1002/rob.21791,6600882,7989421,8206091}. It has been increasingly stressed in the literature that timely and regular monitoring of steel bridges ensures the safety of transportation vehicles. Environmental degradation (e.g., rain, wind, solar radiation), continuous surface-level friction, overloading, and other factors lead to deterioration of different structures on steel bridges. Continuous steel bridge monitoring is necessary to ensure transportation safety and proper maintenance. The tasks can be done manually, however, it is time-consuming, labor intensive, dangerous, affect to the traffic, and sometimes inaccessible for human in complex structures. For the reasons, there are varieties of robotic platforms \cite{Seo_TMECH2013, nguyen2020practical, Wang_TRO2017, Nguyen_IROS2019, la2020steel} developed to support human to do the task. These robots are magnetic-based that help them traverse on multiple angles of steel bridge structures. Most of the robots are controlled manually by an operator.

\begin{figure}[ht]
    \centering
    \setcounter{subfigure}{0}
    \subfigure[]{\includegraphics[height=0.2\linewidth, width=0.25\linewidth]{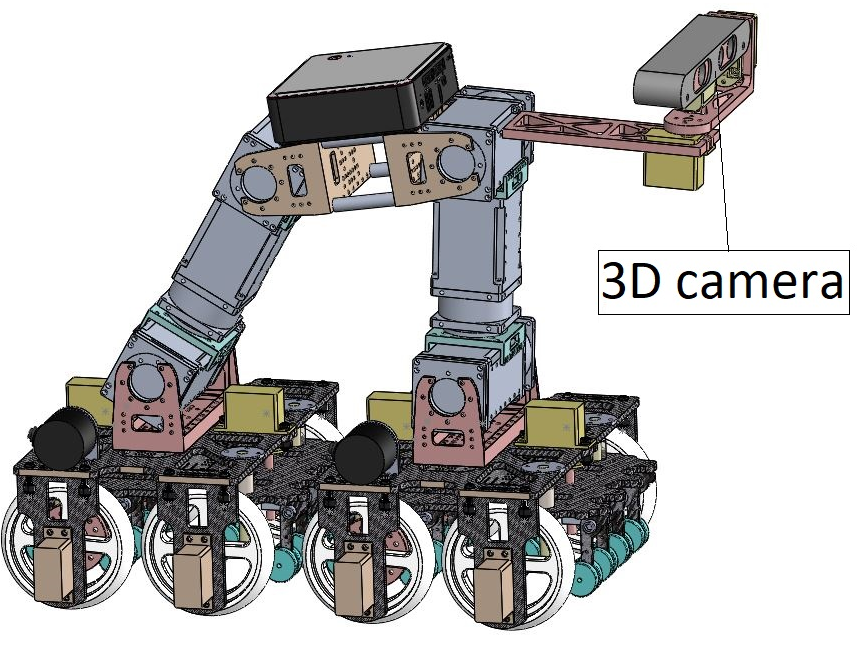}}
    \subfigure[]{\includegraphics[height=0.2\linewidth, width=0.25\linewidth]{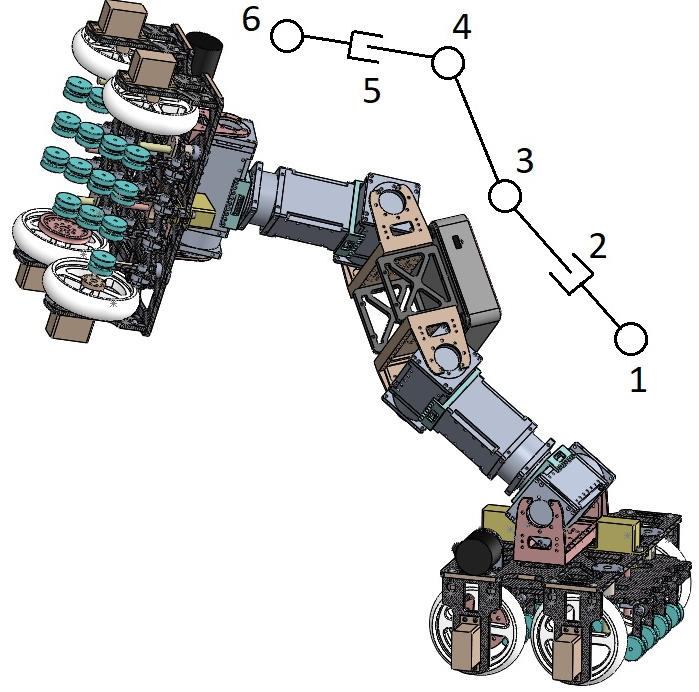}}
    \subfigure[]{\includegraphics[height=0.2\linewidth, width=0.25\linewidth]{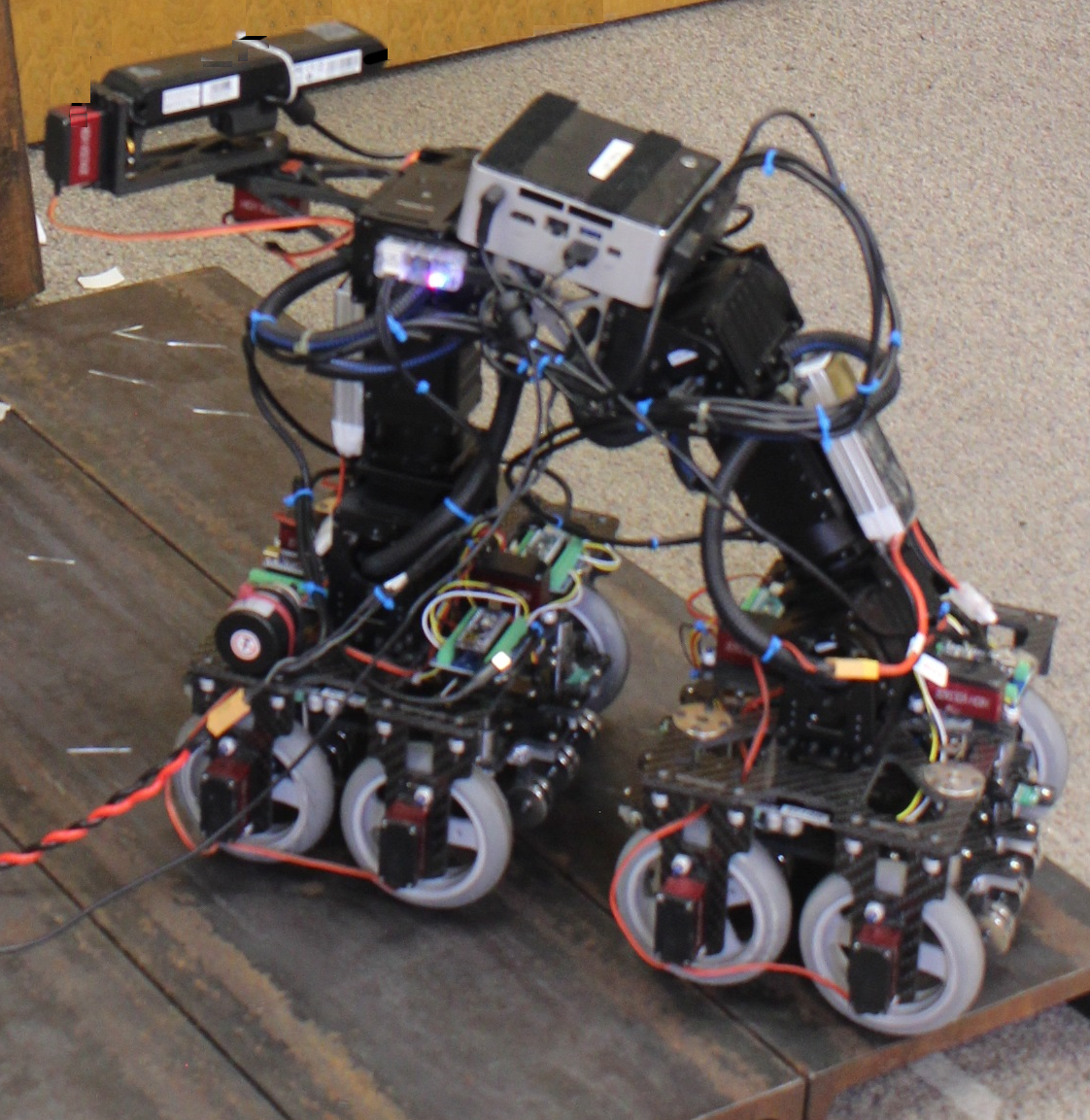}}
    \caption{ARA robot model \cite{nguyen2020practical} in (a) \textit{mobile}, (b) \textit{inch-worm modes}, and (c) real robot}
    \label{fig:newModule1}
    \vspace{-10pt}
\end{figure}

As an effort to go further in the field, the Advanced Robotics and Automation (ARA) Lab of the University of Nevada, Reno has developed a bio-inspired hybrid robot - ARA robot \cite{nguyen2020practical, bui2020control} (Fig. \ref{fig:newModule1}) with the aim to inspect a steel bridge structure autonomously. The robot is able to work in two modes: (1) mobile to traverse on smooth steel surface, and (2) inchworm - to change/jump to another steel bar surface. In this research, a control and navigation framework is proposed for the ARA Lab's robot to move easily and autonomously on smooth steel surfaces, and jump to another steel surfaces, which are adjacent or higher than the current one.

\begin{figure}[ht]
    \centering
    \setcounter{subfigure}{0}
    \subfigure[]{\includegraphics[height=0.2\linewidth, width=0.38\linewidth]{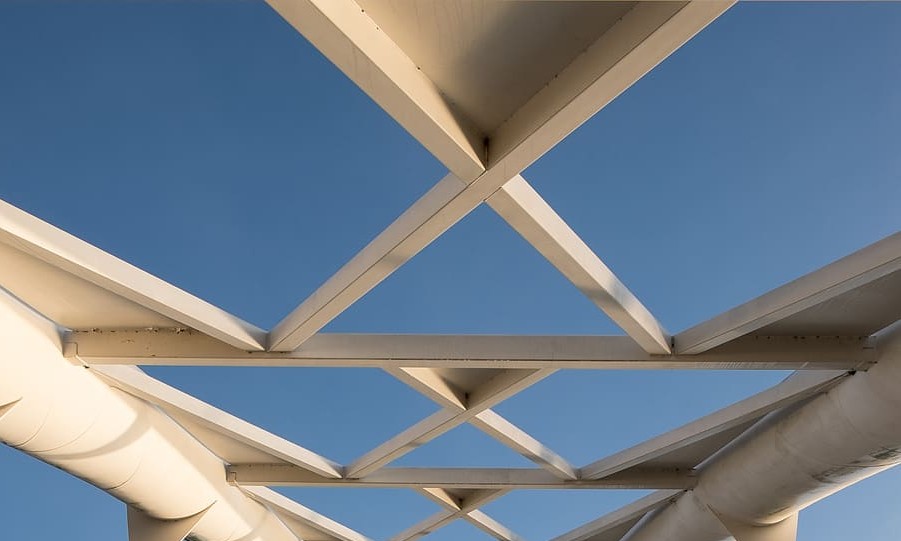}}
    \subfigure[]{\includegraphics[height=0.2\linewidth, width=0.38\linewidth]{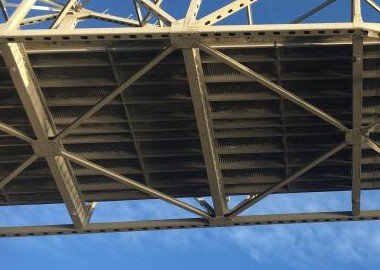}} \\
    \subfigure[]{\includegraphics[height=0.2\linewidth, width=0.38\linewidth]{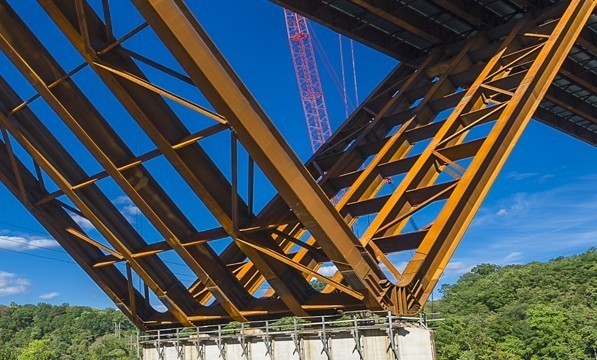}}
    \subfigure[]{\includegraphics[height=0.2\linewidth, width=0.38\linewidth]{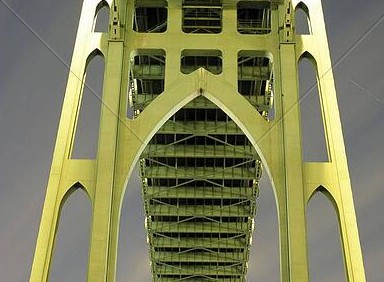}}
    \caption{The typical steel bridges structure: (a) cross shape, (b) K-Shape, (c) L-Shape, (d) and combination of shapes}
    \label{fig:TypeStruct}
\end{figure}

To move on smooth steel surfaces, the robot needs to navigate itself on varieties of structures in steel bridge as shown in Fig. \ref{fig:TypeStruct}, which consists of popular structures as \textit{Cross-}, \textit{T-}, \textit{I-}, \textit{K-} and \textit{L-} shape. The structure's complexity and varied dimensions make motion planning task is very challenging, requires the robot's perception about the structure and a method that is able to make use of limited work space. Moreover, to navigate the robot to inspect a bridge continuously, the navigation system needs to build a path in large scale. Combining with a depth sensor to collect data in a far distance, a method is developed to build and represent a steel bridge structure as a graph. Moreover, we propose a variant of open Chinese Postman Problem (VOCPP), which determines the shortest path to inspect all available steel bars on the structure with difference of starting and ending points.

\subsection{Literature Review}
 Most of steel bridges are monitored by civil inspectors manually \cite{NguyenLa_IROS2019}. However, due to the complex structural composition and inaccessible regions of the bridges (e.g., pipes, poles, overhead cables), the manual inspection of these regions is a perilous task for human inspectors. Additionally, manual inspection is time-consuming, labor-intensive, and disruptive to traffic. It is for this reason that different robotic solutions have been developed for automated steel bridge inspection \cite{Pham_ICERA2020,Nguyen_SHMII2019, Seo_TMECH2013, La_Robotica_2019, Wang_TRO2017, PL_Allerton2016, Kamdar2015, PL_ISARC2016, MaggHD}. These robots are equipped with different adhesion mechanism (e.g., magnetic wheels, pneumatic, suction cups, bio-inspired grippers), visual sensors (e.g., monocular, stereo vision, RGB-D sensors) and other sensory modalities to facilitate navigation and inspection (e.g., IMUs, eddy current sensors) \cite{Pham_ICERA2020, La_Robotica_2019, Wang_TRO2017, Kamdar2015, PL_Allerton2016,PL_ISARC2016, Seo_TMECH2013}. Adhesion force generated by either permanent or electro magnets attached on the robot enables them \cite{MaggHD} to adhere and navigate flexibly on smooth steel surfaces. The incorporation of legged mechanism with electromagnets allows robots to assist in locomotion and traversal through complex steel structures \cite{Magnapods}. These robots are designed for a particular environment, lacking the deploy-ability in many unstructured environments. A flexible and versatile climbing robot was designed in \cite{NguyenLa_IROS2019}, which was equipped with 5-DOF arm, eddy current sensor and RGB-D sensors for inspection of steel bridge surfaces, especially for inaccessible regions of the bridges. Another type of climbing robot was developed by \cite{Versatrax} with untouched magnet blocks to move efficiently on metal surfaces. Although these robots alleviated the difficulty of moving on complex steel surfaces, they were controlled manually by cables or remote instruction from human operators.

There is a number of work related to the navigation of inspection robots for steel bridges  \cite{kotay1996navigating, xie2011edge, pagano2017approach}, which helps the robots move in a local area, and assume the robot dimension quite small comparing to the workspace.
In \cite{kotay1996navigating} particularly for autonomous steel bridge inspection robot, the authors proposed a task-level primitive and online navigation combining with IR sensor, which helps the robot move in local area. In \cite{xie2011edge}, the authors proposed a method to detect edges on a large surface, which is used in navigation. The method in \cite{pagano2017approach} supported the small size robot to move in a large-inside steel bridge space. Their navigation work here assumed that the robot's dimensions are quite small comparing to the workspace. With the limit of steel bar dimension, our research needs to handle a new circumstance for robot navigation and motion planning.

There is an important feature in navigation for steel bridge inspection robots: the dimensions of steel bars are limited, and the robots have small space to make a motion. The methods in \cite{deits2015computing, jatsun2017footstep, hildebrandt2019versatile} are able to build very nice convex regions, which make the construction of configuration space easy. The approximation methods, however, reduce the dimension of the workspace, and could make the robot motion infeasible. To overcome these problems, in this paper we propose a method, which segments the workspace into multiple clusters and represents it by a set of boundary points. 
The method can use all the possible area in the workspace, thus increases the probability of finding a path for the robot. With the irregular shape of the steel bar, Expectation Maximization - Gaussian Mixture Model (EM-GMM) method \cite{reynolds2009gaussian, pfeifer2019expectation, nguyen2012fast, blekas2005spatially} is utilized to segment the data into irregular dimensional clusters.

When perceiving the working space, the navigation system  represents the bridge structure as a graph. Estimation of features from point cloud data receives a significant attention from researchers \cite{goforth2020joint, gandler2020object, kraemer2017simultaneous}. These researches worked on a particular small object to the view space of the depth sensors. In our research, the bridge is large and usually over than the sensor view. A graph construction algorithm is developed for this purpose. From the built graph, the next step is to determine a shortest path to inspect all steel bars in the structure, that is called \textit{inspection route}  or \textit{Chinese Postman  Problem}  (CPP) \cite{edmonds1973matching}. There are several requirements for the shortest path in our context. The real bridge is too long for the depth sensor to collect data in one frame, thus the bridge is separated into multiple parts. As the robot finishes inspection in one part, it will move the next one to continue the task. Therefore, the robot starts at one point and ends in another point. The starting and ending points can be anywhere on the bridge structure, which are convenient for the robot to perform the next task.
There is a number of research working on CPP problem \cite{thimbleby2003directed, eiselt1995arc, ahr2006tabu}, however, there is no work satisfying our requirement of arbitrary starting and ending locations. Therefore, we propose a variant of open CPP (VOCPP) algorithm to handle the problem.

\subsection{Contributions}
Steel bridge inspection is a continuous process, the primary goal of our research is to develop a fully autonomous robotic system to automate this task. In this research, we proposed a control and navigation framework for the ARA robot to navigate autonomously on steel bridge structures. 
The contributions of this paper are then as follows:
\begin{itemize}
    \item A control framework to help the ARA robot switch between two operation modes (mobile, inch-worm) autonomously. The switching control determines the availability of the planar surface, its area and height to decide the next transition;
    \item A non-convex boundary estimation algorithm to find the boundary of the steel bar point cloud, that utilizes the availability of limited steel bar surface.;
    \item An area estimation algorithm using point cloud data from RGB-D sensor to allow the robot to assess area availability for transitioning from one plane to another. This algorithm determines if the available area is sufficient for the robot's foot transition;
    \item An efficient algorithm to segment the steel bridge structure into steel bars and cross area regardless any kinds of input structure (tested on \textit{T-}, \textit{K-}, \textit{I-}, \textit{L-} and \textit{Cross-} shape);
    \item An algorithm to construct an undirected graph from the steel bridge structure data;
    \item \textit{VOCPP} algorithm to find the shortest path for the robot to inspect all steel bars in the graph with difference between the starting and ending points;
    \item A method to determine whether a robot configuration belongs to free configuration, with the input as a set of cluster boundary of steel bridge structure.
\end{itemize}

The organization of the remainder of this paper is as follows. Section~\ref{sec_control} introduces the control framework of the ARA robot. Section ~\ref{sec_navi} presents the navigation of the ARA robot in mobile transformation. Section~\ref{sec_results} discusses the experiment setup and results. Lastly, Section \ref{sec_conclusion} covers the conclusions with analysis and potential future development.

\section{Control Framework}\label{sec_control}
\begin{figure}[ht]
\centerline{\includegraphics[width=0.65\linewidth]{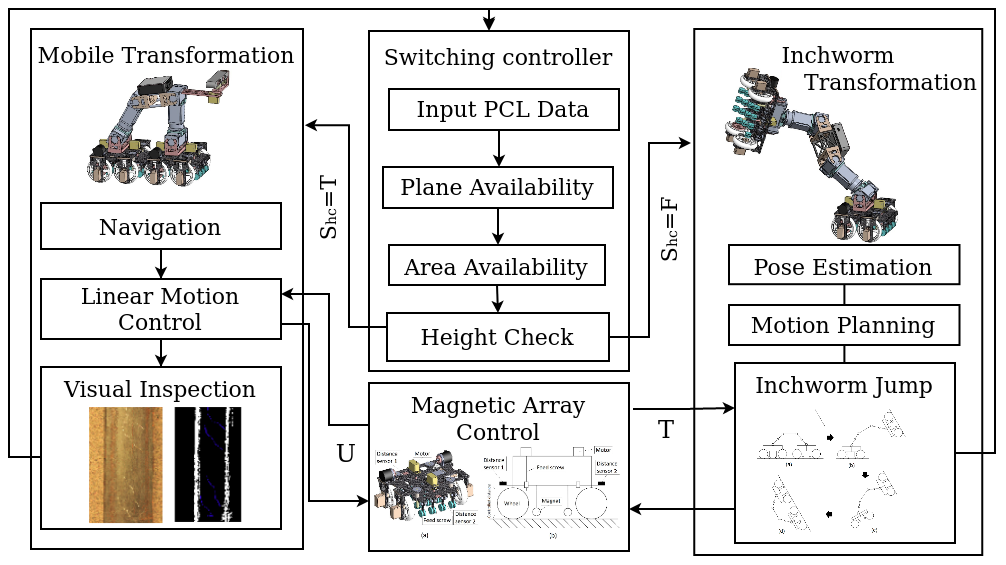}}
    \caption{The proposed control system framework for autonomous navigation}
    \label{fig:OverallFramework}
\end{figure}

\begin{figure}[ht]
\centering
\setcounter{subfigure}{0}
\centerline{\includegraphics[width=0.55\linewidth]{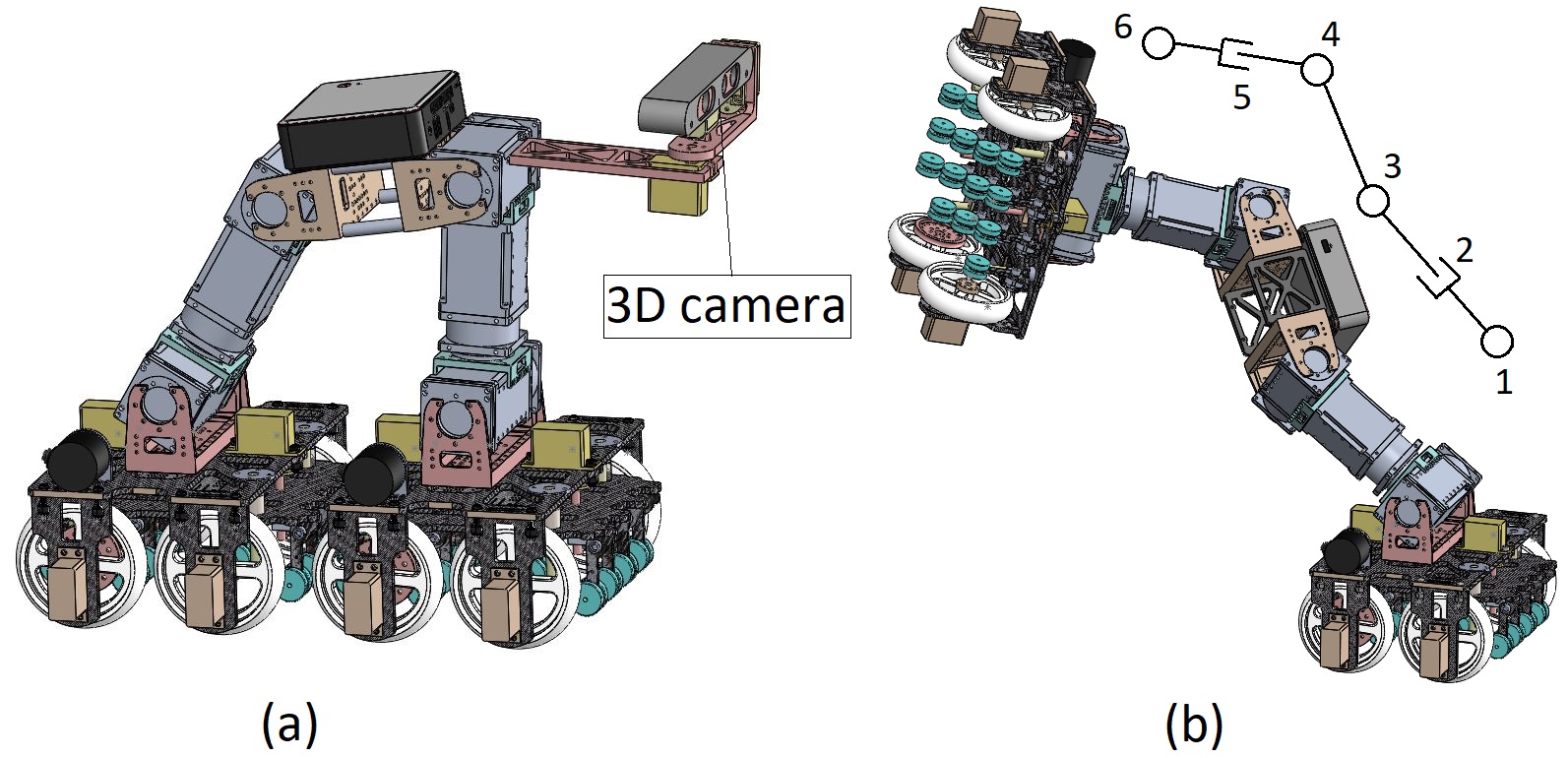}}
\caption{ARA robot in (a) \textit{mobile} and (b) \textit{inch-worm transformation}}
\label{fig:newModule}
\end{figure}

ARA robot can configure itself into two transformations: mobile and inch-worm. In this work, we integrated a switching control mechanism (shown in Fig. \ref{fig:OverallFramework}) to the robot \cite{bui2020control}. This control mechanism enables the robot to change its transformations depending on environmental conditions. When traversing on continuous and smooth steel surfaces, the robot activates the \textit{mobile transformation} as shown in Fig. \ref{fig:newModule}(a). The robot traverses on steel bridge structures by using a navigation framework which is presented in detail in section \ref{sec_navi}. The path outputted from the navigation framework is used to control the robot motion. Steel bridge visual inspection is also implemented on this transformation. Moreover, the robot can move on an inclined steel surface by the adhesion forces supported by two magnetic arrays mounted on each robot foot. 
There are two working modes of the magnetic arrays: \textit{touched} and \textit{untouched}, indicating the distance from the magnetic arrays to the steel surface where the robot feet lies on. \textit{Touched mode} means the distance is zero, and the \textit{untouched} one keeps the distance around $1 mm$. The \textit{mobile transformation} requires both magnetic arrays operating in untouched mode to generate two magnetic adhesion forces, which are enough for the robot standing on the inclined surface, and in same time still let the robot can move by its wheels. The robot switches into \textit{inch-worm transformation} (Fig. \ref{fig:newModule}(b)) when it detects a complex steel surface and cannot move on wheels, then activates an inch-worm jump to the next surface. As performing inch-worm jump, only one of the robots feet touches the steel surface. To create enough adhesive force for the robot standing, the magnetic array is switched to \textit{touched mode}, which fully allows this array to adhere the steel surface. The switching control mechanism controls the movement of the robot, detects environment type and sends the appropriate command to executable nodes.

\begin{figure}[ht]
\centerline{\includegraphics[width=0.45\linewidth]{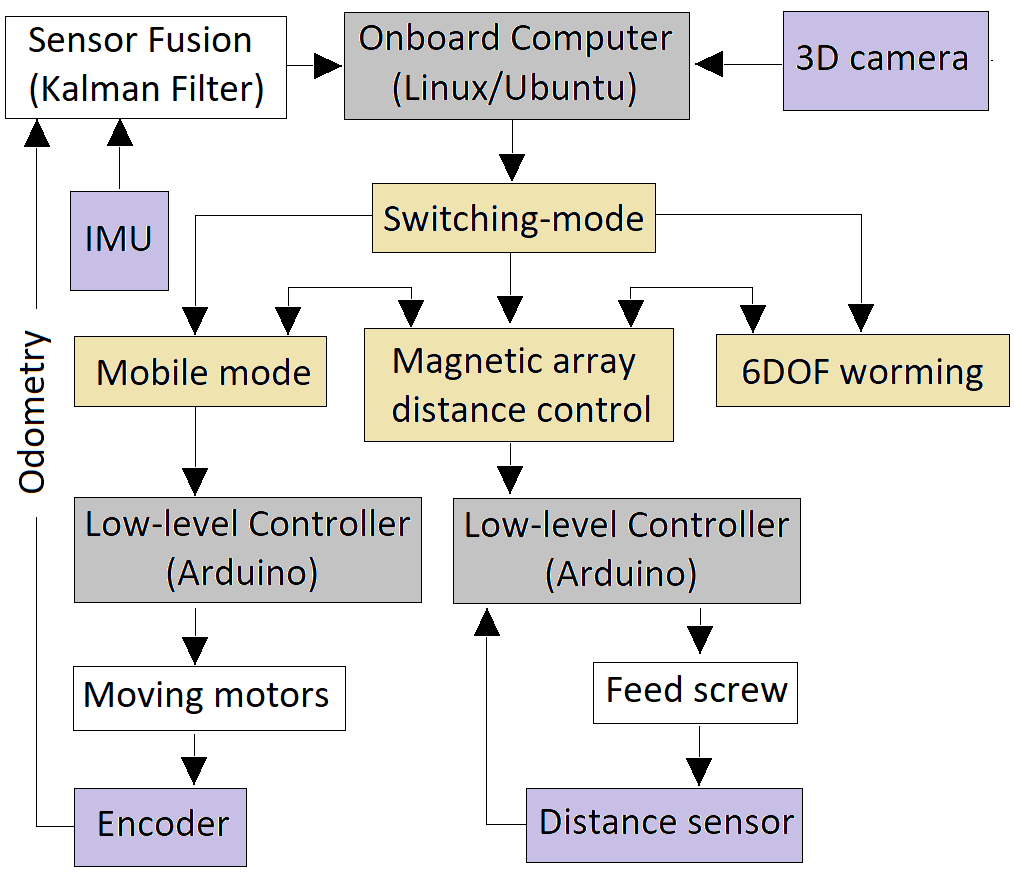}}
    \caption{The control architecture integrated into ARA robot \cite{nguyen2020practical} }
    \label{fig:Control_Structure}
\end{figure}

The control architecture of ARA robot is comprised of multiple low-level and high-level control structures \cite{bui2020control}. Several tasks are performed by the low-level control structure (Arduino). The wheel's velocity, encoder reading, and the magnetic array function are performed in this control level. The high-level control embedded in an on-board processor manages switching control function, point cloud data processing, inverse kinematics and motion planning. The arrangement of the high level and low-level controls is shown in Fig. \ref{fig:Control_Structure}.

The ARA robot's control and navigation framework consists of four modules: \textit{switching control}, \textit{inch-worm transformation}, \textit{mobile transformation}, and \textit{magnetic array control}. An overview of the overall framework is shown in Fig. \ref{fig:OverallFramework}. The first two modules and the \textit{Navigation} in the \textit{mobile transformation} module are the contributions of the author's work that will be presented in detail. The \textit{Navigation} part is introduced separately in Section \ref{sec_navi}.

\subsection{Switching Control}
The switching control $S$ enables the robot to autonomously configure itself into two transformations (mobile and inch-worm). The control employs switching function $S$, represented in Eq.\ref{eq:switchingFunction}. The function takes as input three Boolean parameters: plane availability $S_{pa}$, area availability $S_{am}$ and height availability $S_{hc}$. These parameters determine if there is any still surface available, while enabling the estimation of the area of the surface and its height. A logical operation is performed on these parameters using function $f(.)$. The function's parameters are estimated from 3D point cloud data of steel surface.
\begin{equation} \label{eq:switchingFunction}
S = f(S_{pa}, S_{am}, S_{hc}) = S_{pa} S_{am} S_{hc}.
\end{equation}

\begin{figure}[ht]
    \centering
    \includegraphics[width=0.45\linewidth]{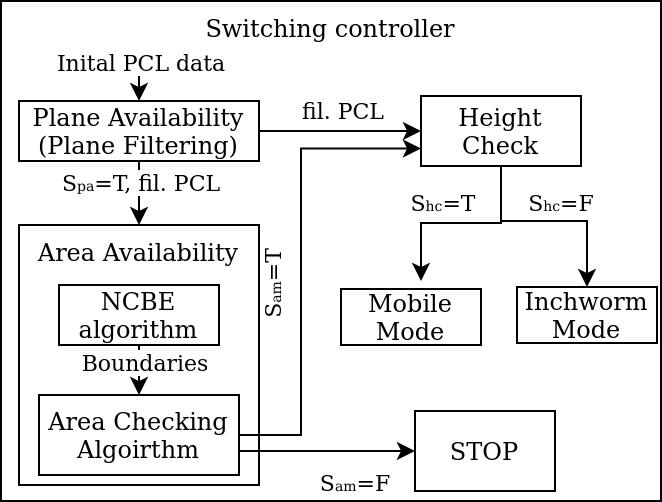}
    \caption{Switching Control's Structure}
    \label{fig:switching_control}
\end{figure}

The \textit{Switching Control}'s structure is presented in Fig. \ref{fig:switching_control}. As receiving a PCL data, it is processed by Plane Availability algorithm, which outputs the plane if it is available. The plane data is then sent to Area Availability check, which consists of two algorithms: NCBE and Area Checking. If the plane is large enough for the robot stepping or moving on, $S_{am} = true$ is transited to the Height Check algorithm. Otherwise, it will stop the robot. 
The robot is configured into \textit{mobile transformation} if $S_{hc}$ returns a true value. The false value configures the robot into the \textit{inch-worm transformation}.

\textbf{Plane availability:} The 3D PCL of a steel surface is processed using \textit{pass-through} filtering, \textit{downsampling}, and \textit{plane detection} \cite{buiIRC2020Sort}. The \textit{plane detection} applied the RANSAC method \cite{strutz2010data} extracts the planar point cloud $P_{cl}$ from the initial point cloud. The plane availability is checked using  Eq. (\ref{Eq:PlaneCheck}):
\begin{equation}\label{Eq:PlaneCheck}
\begin{cases}
    S_{pa} = False, \, if \, P_{cl} = \emptyset \\
    S_{pa} = True, \, otherwise.
\end{cases}
\end{equation}
Moreover, two functions \textit{get\_centroid} and \textit{get\_normal\_vector} provide the point cloud's centroid $C_{P_{cl}}$ and normal vector $\vec{N_{P_{cl}}}$ of the point cloud.

\textbf{Area availability:} The robot feet requires an area estimation of the available planar surface area $P_{cl}$. It is essential to ensure the availability of sufficient area for successful robot transition. This is a popular problem in legged-robot, which has been investigated carefully in \cite{deits2015computing, jatsun2017footstep, hildebrandt2019versatile}. In \cite{deits2015computing, jatsun2017footstep}, the authors proposed the convex-based algorithms, which deployed convex optimization problem to determine an obstacle-free ellipsoid (convex one), then estimate step-able areas for a biped robot. In \cite{hildebrandt2019versatile}, the authors proposed an algorithm to determine the valid convex collision-free regions with geometrical constraints of obstacles. In those algorithms, a portion of the step-able area, especially as the vertex number is small, was not considered due to the convex approximation. It is a problem for our inspection robot with large feet pair due to limited step-able areas on steel bridges. Those algorithms may not be possible to find a step-able area for the ARA robot to worm in many cases.
Thus, we developed two algorithms, which can process a non-convex plane boundary efficiently to estimate a step-able area for the ARA robot. \textit{Algorithm \ref{alg:boundaryestimation}} extracts a non-convex boundary from the planar point cloud, then the second one - \textit{Algorithm \ref{alg:areaestimation}} checks the sufficiency of the available planar surface. 

\begin{figure}[ht]
\centerline{\includegraphics[width=0.55\linewidth]{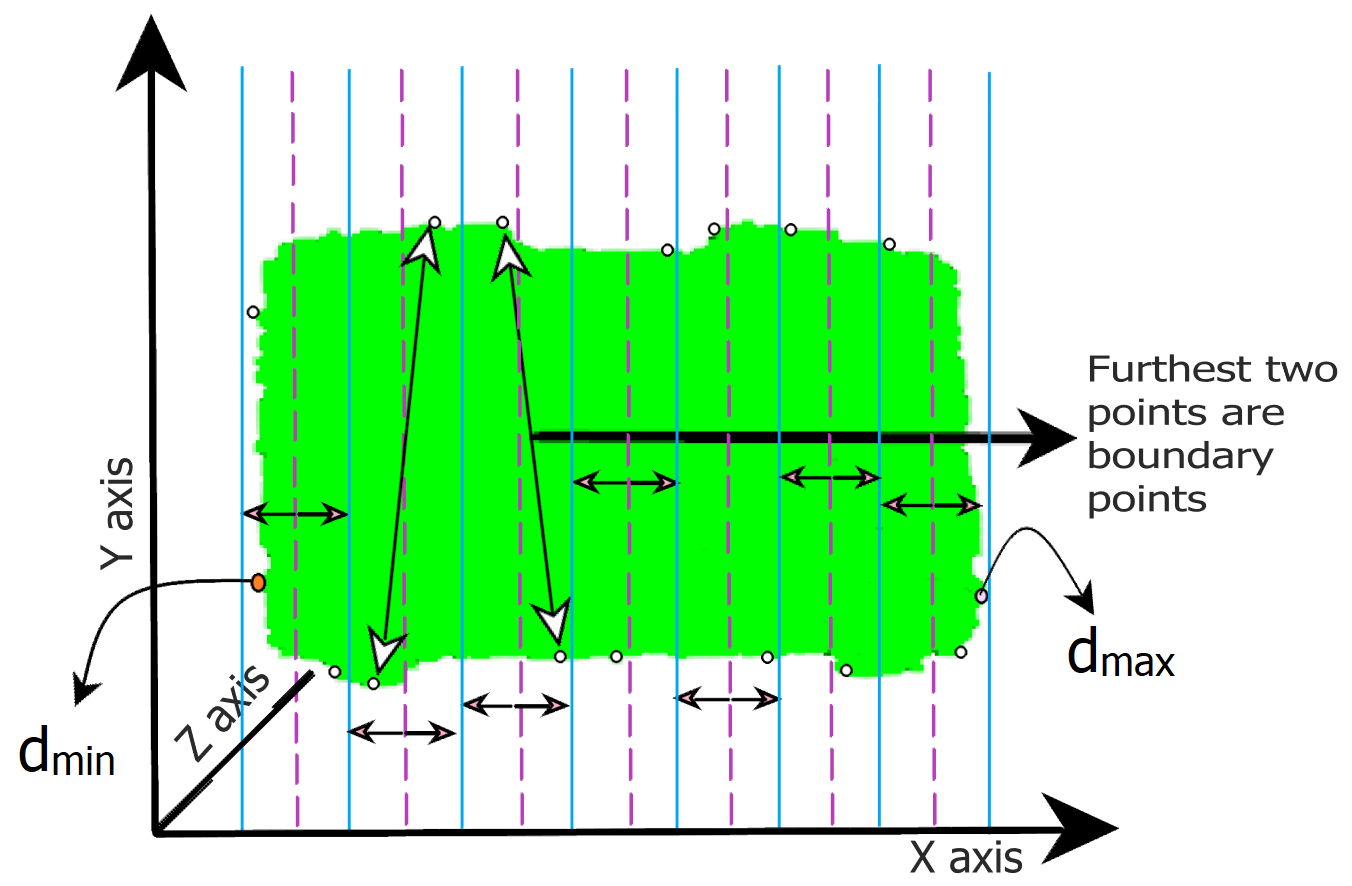}}
    \caption{ Boundary point estimation from 3D point cloud data}
    \label{fig:BoundaryAlgorithm}
\end{figure}

\begin{algorithm}[ht]
\small
\caption{Non-convex boundary point estimation}\label{alg:boundaryestimation}
    \begin{algorithmic}[1]
        \Procedure{BoundaryEstimation}{$P_{cl}, \alpha_s$}
            \State $Planes$ = \{$xy, yz, zx$\}
            \State $d_{min}$ = $\forall_{i \in Planes}$ {\ssmall //Point along minimum value of plane $i$}
            \State $d_{max}$ = $\forall_{i \in Planes}$ {\ssmall //Point along maximum value of plane $i$}
            \State Initialize $B_s = \{\}$
            \For{$p \in Planes $}
                \State $i \to 1$
                \While {$sl_{p_i} < d_{max}$}
                    \State $sl_{p_i} = d_{min_p} + i*\alpha_s$
                    \State $PS_{p_i}$ = Set of points in range  $sl_{p_i}\pm \alpha_s/2$
                \State { $P_{cl_{A}}, P_{cl_{B}} = \underset{\forall\{P_i,P_j\} \in PS_{p_i}}{\mathrm{argmax}} \{ \left\Vert{P_i-P_j}\right\Vert$}\}
                    \State $B_s = B_s \cup \{P_{cl_{A}}, P_{cl_{B}}\}$ 
                    \State $i = i+ 1 $
                \EndWhile
            \EndFor \\
            \Return $B_s$
        \EndProcedure
    \end{algorithmic}
\end{algorithm}
The boundary points estimated in \textit{Algorithm \ref{alg:boundaryestimation}} are performed by a window-based approach. The algorithm's input is the point cloud $P_{cl}$ of the estimated planar surface and a slicing parameter $\alpha_s$. At first, we calculate the two furthest points represented as $d_{min}$ and $d_{max}$ in the point cloud $P_{cl}$ along each plane, then the point cloud is divided into multiple smaller slices along the three planes. For each slice in a particular plane $p$, the slicing index  $sl_{p}$ is determined, which represents the center coordinate of the slice as shown in line $9$ of algorithm \ref{alg:boundaryestimation}. After that, the point sets $PS_{p}$ in the range $sl_{p}\pm \alpha_s/2$ is extracted from $P_{cl}$. This sliding factor is experimentally determined based on the point cloud size. 
For each set of points from $PS_{p}$, two furthest points ($P_{cl_{A}}, P_{cl_{B}}$) are extracted. These two points are estimated as the boundary point for that particular slice and added to the boundary point sets $B_s$. This approach helps the algorithm works well with plane with small holes inside. A pictorial representation of the boundary estimation algorithm is shown in Fig. \ref{fig:BoundaryAlgorithm}.

\begin{algorithm}[ht]
\small
\caption{Area Checking and Pose Estimation}\label{alg:areaestimation}
    \begin{algorithmic}[1]
        \Procedure{Area}{$B_{s}, C_{P_{cl}}, \vec{n_{P_{cl}}}, w, l, t, S_{am}$}
            \State $N_{clos}$ = Find $n$ closest points to $C_{P_{cl}}$ from $B_{s}$
            \For { $N_i \in N_{clos}$ }
                \State $R = \{\}$, {\ssmall //Estimated rectangle corner points}
                \State $\vec{e_{x_i}} = N_i - C_{P_{cl}} $
                \State $\vec{e_{z_i}} = \vec{n_{P_{cl}}}$
                \State $\vec{e_{y_i}} = \vec{e_{x_i}} \times \vec{e_{y_i}} $
                \State $k_w = \frac{w}{|\vec{e_{x_i}}|} e_{y_i} $ and $k_l = \frac{b}{|\vec{e_{y_i}}|} e_{x_i}$,
                \State $\{R_1, R_2\} = \{ N_i + k_w, N_i - k_w \}$
                \State $R = R \cup \{R_1, R_2\} $
                \State $R = R \cup \{R_1+k_l, R_2+k_l\}$
                \State $M = \forall_{r_i \in R} \{ \frac{r_i + r_{i+1}}{2}\}$
                \State $R = R \cup M$
                \State $S_{am} = True$
                \For { $r_i \in R$ }
                    \State $Q_i$ = Find $m$ closest points to $r_i$ 
                    \State $d_{r_i} = \left\Vert{d_{r_i}, C_{P_{cl}}}\right\Vert$ and $d_{Q_i} = \left\Vert{Q_i, C_{P_{cl}}}\right\Vert$
                    \State $S_i = (d_{r_i} < d_{Q_i}) \lor(\frac{d_{r_i}-d_{Q_i}}{{d_{r_{i}}}} <t$
                    \State $S_{am} = S_{am} \land S_i$ 
                \EndFor
                \State Pose = \{Orientation, Position\}
                \If{ $S_{am} == $ True}
                    \State $R_c$ = Centroid of $R$
                    \State Orientation = ($\vec{e_{x_i}}, \vec{e_{y_i}},  \vec{e_{z_i}}$)
                    \State Position = $(x_{R_c}, y_{R_c}-l/4, z_{R_c})$
                    \State \Return Pose
                \EndIf
            \EndFor
            \State \Return False 
        \EndProcedure
    \end{algorithmic}
\end{algorithm}

After determining the boundary points $B_s$, the area availability variable $S_{am}$ is estimated by using \textit{Algorithm \ref{alg:areaestimation}}. 
The inputs are the boundary points $B_s$, point cloud centroid $C_{P_{cl}}$, normal vector of point cloud $\vec{n_{P_{cl}}}$, length $l$ and width $w$ of robot feet,  and distance tolerance $t$. At first we calculate the $n$ closest points ($N_{clos}$) from $B_s$ to  the point cloud centroid $C_{P_{cl}}$.  For each point, $N_i$ in the set $N_{clos}$, a set of computations is performed to estimate the plane corners for adherence to robot wheels. The coordinate frame vectors $\vec{e_{x_i}}, \vec{e_{y_i}}$ and $\vec{e_{z_i}}$ are calculated for point $N_i$. In the next step, the algorithm estimates the corner points of a rectangle of width $w$ and length $l$, which is also robot foot's width and length, respectively. We estimate the rectangle's edges parallel along the vectors $\vec{e_{x_i}}$ and $\vec{e_{y_i}}$. Therefore, the four corners $R$ of the rectangle are estimated using these two vectors. Additionally, we include four middle points of the estimated rectangle corners in $R$ to alleviate point cloud collection error as well as accommodate for the non-convex shape of the steel surface. After the estimation step, we find $m$ closest points to $R$ from the $B_s$ to measure if the points in $R$ are inside the boundary. Hence, we calculate the distance from point cloud centroid $C_{P_{cl}}$ to $R$ and $Q$, respectively. The algorithm considers a point, which lies inside the boundary if its tolerance is less than $t$ and the distance to centroid should be less than its neighbors.
The algorithm's performance is presented in Fig. \ref{fig:boundaryRectangle}(b)-(d). When the value of $S_{am}$ is true for all the conditions, we consider those sets of points as rectangular points.

\textbf{Height availability:} The height availability $S_{hc}$ is crucial for the switching control. Based on this parameter, the switching control activates the robot transformations. At first the point cloud's centroid $C_{P_{cl}}$ is calculated along the camera frame $f_c$. Then it is transformed to the robot base frame $f_{rb}$ using Eq. \ref{Eq:transMatrix}.
\begin{equation}\label{Eq:transMatrix}
    P_{C_{f_{rb}}} = T_{f_{rb}f_c} P_{C_{f_c}},
\end{equation}
where $(P_{C_{f_{rb}}}$, $P_{C_{f_c}})$ are coordinates of the centroid $C_c$ in the camera frame and the robot base frame, respectively.  $T_{f_{rb}f_c}$ is the transformation matrix from the camera frame $f_c$ to the robot base frame $f_{rb}$.

The plane height $z_{f_{rb}}$ coordinate is then compared to the robot base height. If they are equal, the returned result is \textit{true}, and the robot is configured as \textit{mobile transformation}. Otherwise, it returns \textit{false}, and the robot go to \textit{\textit{inch-worm transformation}}. The height availability condition is shown as in Eq. \ref{Eq:heightCheck}.
\begin{equation}\label{Eq:heightCheck}
\begin{cases}
    S_{hc} = True, \, if \,{z_{f_{rb}}} = z_{robotbase}\\
    S_{hc} = False, \, otherwise.
\end{cases}
\end{equation}

\subsection{Inchworm transformation}\label{worming mode}
\begin{figure}[ht]
\centerline{\includegraphics[width=0.4\linewidth]{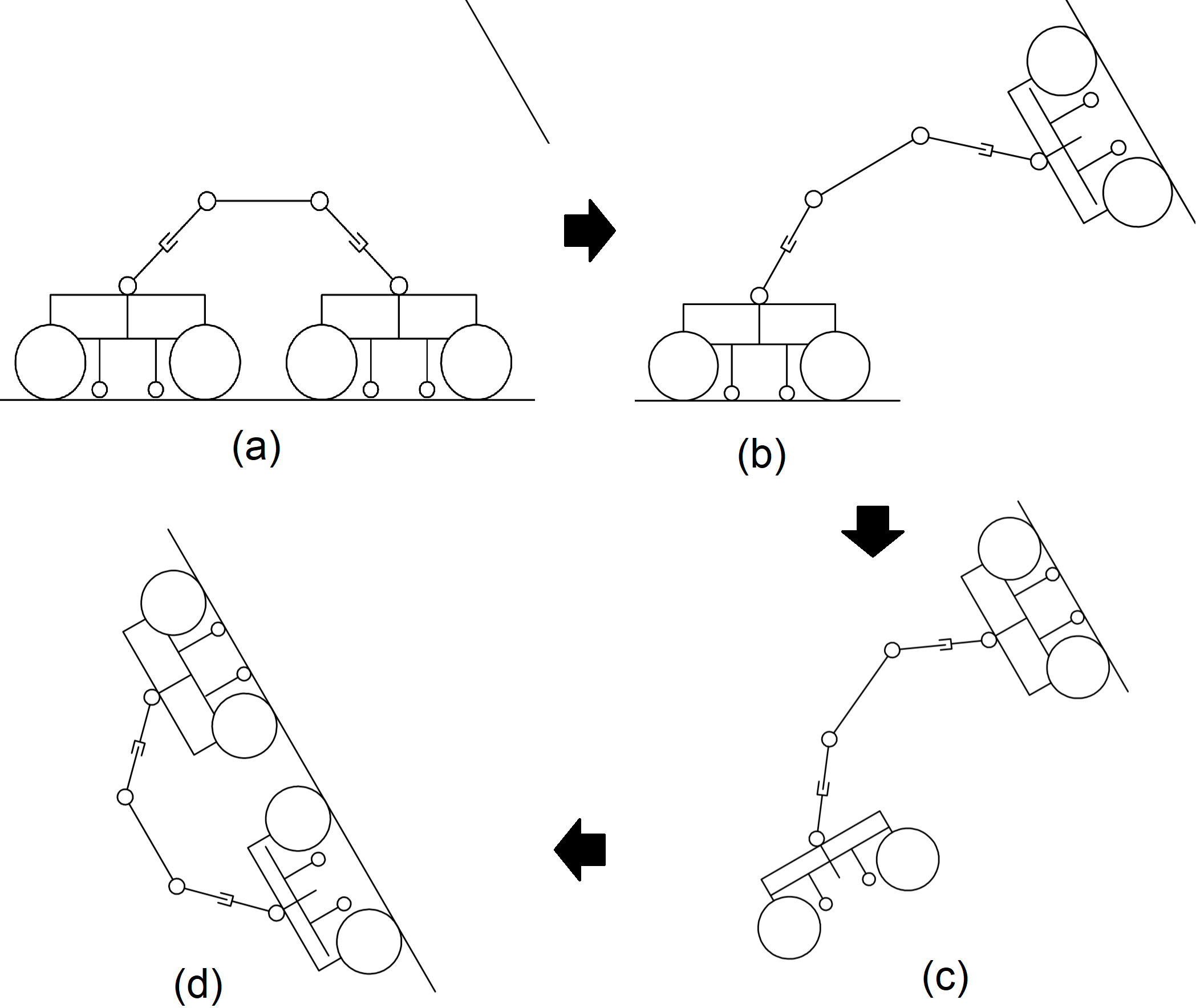}}
\caption{Inch-worm jump from one steel surface to another}
\label{fig:inchWormJump}
\end{figure}

The inch-worm transformation enables the robot to perform an inch-worm jump from one steel surface to another as shown in Fig. \ref{fig:inchWormJump}. At first, the permanent magnet array on the second foot of the robot is set to \textit{touched mode}, which adheres the leg on steel surface and generates a strong adhesive force for the robot to stand and perform the worming. A controller manipulates the joints to move the first robot leg towards the target plane as shown in  Fig. \ref{fig:inchWormJump}(b). As the first leg touches the target surface, the first and second permanent magnetic arrays are switched to \textit{touched mode} and \textit{untouched mode}, respectively. After that, the second leg detaches from the starting surface as in Fig. \ref{fig:inchWormJump}(c). Finally, in Fig. \ref{fig:inchWormJump}(d) the second leg is adhered the target surface.

Converting from \textit{mobile} (Fig. \ref{fig:newModule}(a)) to \textit{inch-worm transformation} is challenging for the motion planner to create a trajectory. To have a better performance, a convenient robot pose $P_{conv}$ is proposed where the robot should move to firstly as starting of \textit{inch-worm transformation}. From there, the motion planner will generate a trajectory for the first leg to move the destination. The worming is completed by moving the second leg to the target surface and reform \textit{mobile configuration}. 
To have the target pose, the robot needs to determine the target plane and its pose, which are the outputs of \textit{Algorithm \ref{alg:areaestimation}}.

The flexibility of the robot in worming is made of the six DoF arm. The revolute joints of the arm in Fig. \ref{fig:newModule}(b) can rotate along three different axes separately. For example, the joint $2$ and joint $5$ in Fig. \ref{fig:newModule}(b) are configured to rotate around \textit{y-axis}. The rest of the joints are positioned to rotate around \textit{z-axis}. This configuration was selected for maintaining symmetry so that the manipulator can move efficiently in both worming and mobile mode. Our previous research states in detail elaboration of the robotic arm in \cite{nguyen2020practical}. 

\section{Navigation Framework}\label{sec_navi}
\begin{figure}[ht]
    \centering
    \includegraphics[height=0.45\linewidth]{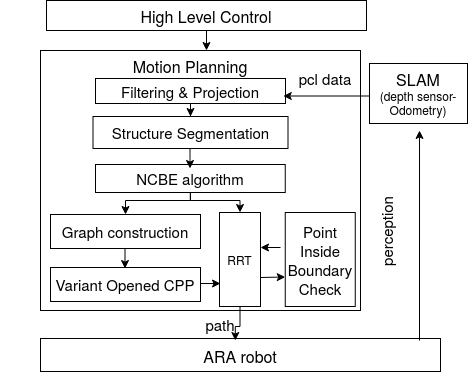}
    \caption{The proposed navigation framework on mobile mode}
    \label{fig:naviFrame}
\end{figure}

As the ARA robot works on \textit{mobile transformation}, it traverses all steel bars of a steel bridge to detect the failure. To do the task in shortest time, it is needed to solve an optimization problem to determine the shortest path, which goes through all the available steel bars at least once \cite{bui2021navigation}. The task is sent to the motion planning module, and the inputs are point cloud (PCL) data, target position, and robot location from SLAM. The PCL data of the steel bridge is filtered and then projected into a 2D plane, which is the \textit{xy} plane of the robot coordinate frame \cite{bui2020control}. The structure segmentation algorithm - \textit{Algorithm 3} separates the processed data into the clusters, then those boundaries are estimated by \textit{Algorithm 2} - Non-Convex Boundary Estimation (NCBE) \cite{bui2020control}. Graph construction algorithm - \textit{Algorithm 4} processes the boundaries data to build a graph that represents the steel bridge structure, then \textit{Algorithm 5} - VOCPP algorithm solves the optimization problem to find the shortest path. The VOCPP algorithm can solve any graph with any starting and ending points.
A single-query path planning such as RRT receives the cluster boundaries and shortest path as inputs and builds a traverse path for the ARA robot to go along the steel bars.
In this path planning, we propose \textit{Algorithm 6} called \textit{Point Inside Boundary Check - PIBC} to determine efficiently the collision and availability of the robot configuration by the boundaries. 
The output of the motion planner is sent to the ARA robot controller, which regulates the robot at the lower level to perform the motion. The framework is shown in Fig. \ref{fig:naviFrame}.

\subsection{Steel Bridge Structure Segmentation}
Motion planning for a robot traversing on a steel bridge is a challenging task because of the limitation of working space (steel bar's dimensions) and the steel bridge structure's diversity. Serious damage occurs if the robot moves out of the steel bar edges. Therefore, steel bridge perception is critical to build a motion planner. The bridge structure can be perceived by a convolutional neutral network (CNN) \cite{narazaki2018automated}, however, it requires a huge data set and expensive tasks such as labeling. Moreover, with the variety of the structures, in the best of the author's knowledge, there is no CNN working on steel bridge structure detection. 

To detect the steel bridge structure, we propose a method based on the bridge's geometric features that they typically consists of two components: (1) steel bars and (2) cross areas. As the method segments the structure into these two components, the bridge structure is represented by a graph whose edges and vertices are the steel bars and cross areas, respectively. 
The steel bars are irregular in shape with one dimension being much longer than other, and their two ends connect two cross areas. The cross areas are regular in shape and its dimensions are similar, and there are at least two steel bars connected with it. The difference in the geometric features of the steel bars and cross areas are used to classify them. 

From the feature analysis, an algorithm is developed based on EM-GMM classification  \cite{reynolds2009gaussian, nguyen2012fast, blekas2005spatially}. The EM-GMM classification is able to separate the structure into multiple clusters with irregular shapes. However, it requires a specific cluster number as input, and if there are several types of distribution in the data, EM-GMM does not works well. To deal with unknown cluster numbers, the algorithm works on a set of cluster numbers and determines which is the most suitable number $n_o$ for cluster segmentation. The number $n_o$ is then inputted into the EM-GMM algorithm \cite{reynolds2009gaussian, nguyen2012fast} that generates a set of clusters. 
\begin{figure}[ht]
    \centering
    \setcounter{subfigure}{0}
    \subfigure[]{\includegraphics[height=0.25\linewidth, width=0.4\linewidth]{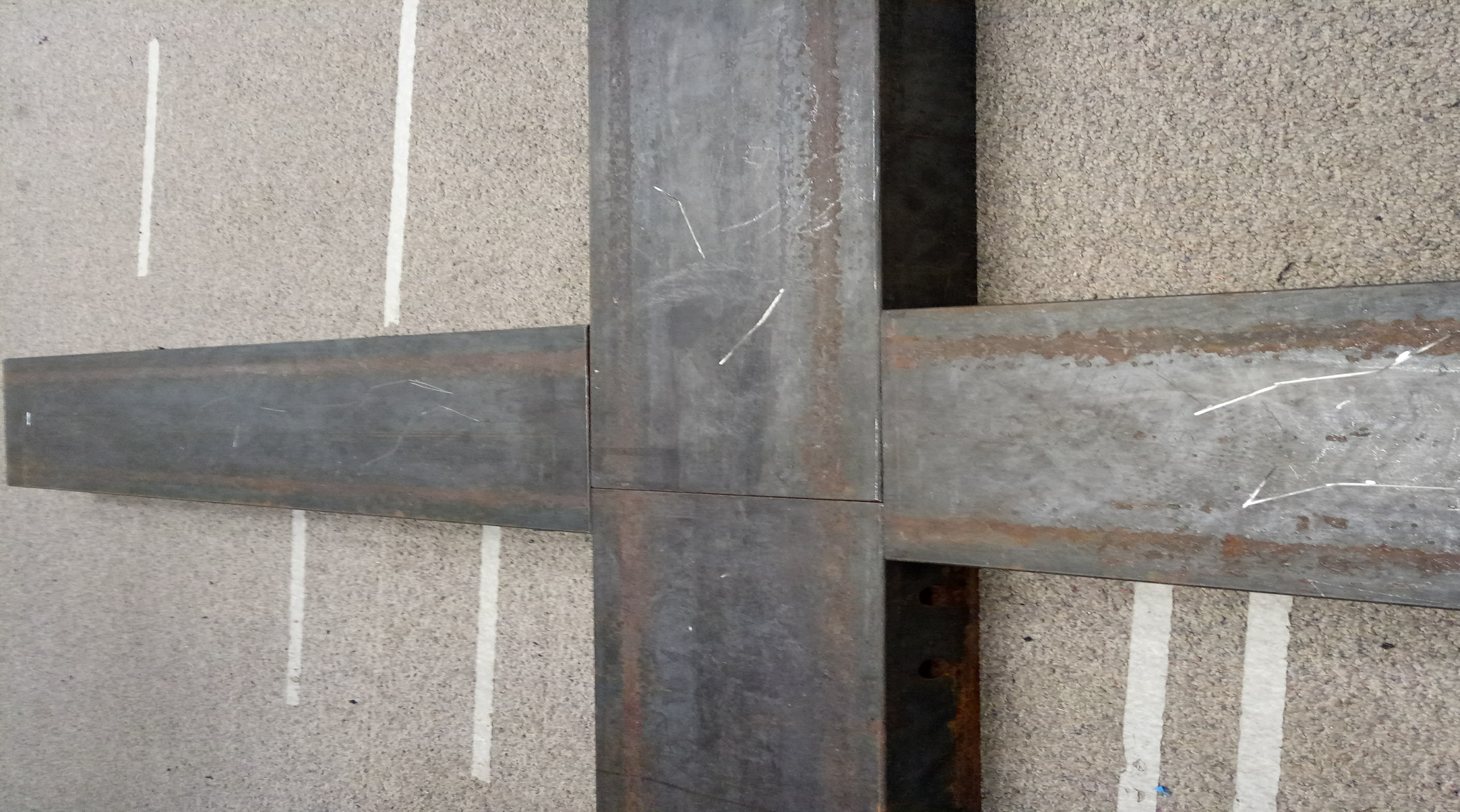}}
    \subfigure[]{\includegraphics[height=0.25\linewidth]{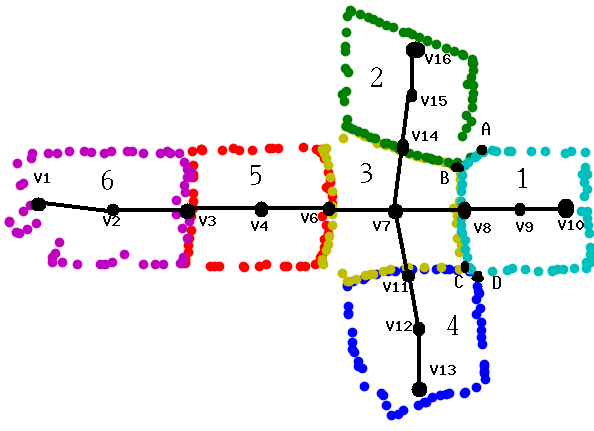}}
    \caption{(a) A \textit{Cross-} shape steel bar (camera view on the right) and (b) its segmentation}
    \label{fig:bar_cross}
\end{figure}

To find the best cluster number for the steel bridge segmentation, a new concept - neighbor cluster - is introduced.  \textit{Two clusters are considered as neighbors if they share a border with the length at least $l_b$}. From that, in Fig. \ref{fig:bar_cross}b, cluster 1 and 3 are considered neighbors because the border \textit{BC} is longer than the threshold $l_b$. Although there are some contacting points, there can be no neighbor-relationship between clusters 1 and 2 because the border \textit{AB} is shorter than $l_b$. The same applies to cluster 1 and 4 since \textit{CD} $\leq l_b$. From this idea, if the cluster numbers $n_o$ are optimal, the cross area should be the one with most neighbor clusters $n_m$ as shown in Fig. \ref{fig:bar_cross}b (cluster 3). Therefore, the first idea is that the cluster number $n_o$ is optimal if it makes the cross area cluster having $n_m$ with highest value.

The most neighbor cluster numbers $n_m$ works well for \textit{K-}, \textit{T-}, \textit{Cross-} shape. However, for the structures such as \textit{I-} and \textit{L-} shapes, the highest number of neighbor is not enough to segment correctly. For \textit{L-}shape, as the cluster number is more than three, there is several clusters with the same neighbor cluster number, and it is not possible to find the cross area cluster. For \textit{I-}shape, there is two cross-area cluster in the structure. Thus, it is needed another feature to handle these shapes, and the difference between the most neighbor number $n_m$ and the second most neighbor number $n_s$ with the same $n_c$ is considered. Combining the two features, the ratio $r$ is defined in Eq. (\ref{Eq:optclus}), and the highest $r$ will be selected. 
\begin{align} \label{Eq:optclus}
    r = \frac{n_m}{n_m + n_s} + \frac{n_m}{n_c}.
\end{align}
In Eq. (\ref{Eq:optclus}), the first term tends to keep $n_c$ small, and the second term tends to make $n_m$ large. $n_c$ is the number of clusters.

\setlength{\textfloatsep}{4pt}
\begin{algorithm}[ht]
\small
\caption{Steel Bridge Structure Segmentation}\label{alg:PclSeg}
    \begin{algorithmic}[1]
        \Procedure{Pcl Segmentation}{$P_{cl}, N_{cmin}, N_{cmax}$}
            \State Initialize $n_{index}, r_n=\emptyset$
            \For{$n_c \in \{ N_{cmin}, N_{cmax}\} $}
                \State $S_c$ = \textit{EM-GMM Algorithm}($P_{cl}, n_c$)
                \For {$j \in \{0,n_c\}$} \label{step:ncbest}
                    \State $b[j] = NCBE(S_c[j], sliding\_factor)$
                \EndFor \label{step:ncbeend}
                \State {Determine neighbor clusters for $S_c[j]$} \label{step:neighClus}
                \State{Get the most and second most neighbor numbers $n_m, n_s$}
                \State{Calculate $r$ from Eq. \ref{Eq:optclus}, and add to set $r_n$}
                \State{Add the $n_c$ in set $n_{index}$}
            \EndFor
            \State{Select the highest $r$ in $r_n$, and get $n_o$ from $n_{index}$} \label{step:r}
            \State {$S_b$ = EM-GMM Algorithm($P_{cl}, n_o$)} \\
            \Return $S_b$
        \EndProcedure
    \end{algorithmic}
\end{algorithm}

The procedure detail is presented by the psudo-code in \textit{Algorithm \ref{alg:PclSeg}}. The inputs are the PCL data, and a range of cluster number from $N_{cmin}$ to $N_{cmax}$. As $n_c$ runs in the range of [$N_{cmin}$ to $N_{cmax}$], EM-GMM algorithm segments the PCL data into a set of clusters $S_c$. From line \ref{step:ncbest} to \ref{step:ncbeend}, NCBE algorithm determines the set of boundaries $b$ corresponding to $S_c$. The neighbor number for each cluster is determined in line \ref{step:neighClus}, and from that the most and second most cluster numbers are specified. From $n_c, n_m$, and $n_s$, the ratio $r$ is calculated, then the $r$ and $n_c$ are putted into the sets $r_n$ and $n_{index}$, respectively. Step \ref{step:r} selects the highest $r$, which is inputted into EM-GMM algorithm to get the most appropriate segmented clusters $S_b$.

\subsection{Graph Construction and VOCPP}
From an inspection requirement aspect, the ARA robot should monitor all available steel bars in the bridge structure to check for any failure. As the cross areas and steel bars are considered as nodes and edges, respectively, the inspection requirement is able to be solved by \textit{inspection route problem} or Chinese Postman Problem (CPP) \cite{edmonds1973matching}. 
Therefore, a graph representing the structure is constructed, then the optimization problem is solved to find the shortest inspection route for the robot.
\begin{algorithm}[ht]
\small
\caption{Graph Construction} \label{alg:GraphEsti}
    \begin{algorithmic}[1]
        \Procedure{Graph Construction}{data cluster set $S_{bo}$, threshold $d_{min}$}
            \State Initialize $E = \{\}, V = \{\}$, current vertex $v_c$
            \State{Calculate the center point set $S_c$ of boundary set $S_b$} \label{step:cenPoint2}
            \State{Determine neighbor matrix $M_n$} \label{step:Neighmatrix}
            \State{Determine borders set $S_{bd}$ among the neighbors from $S_c, M_n$}
            \State{Determine the middle point set $S_{md}$ of $S_{bd}$} \label{step:middlepoint}
            \State{Add all vertices $S_c, S_{md}$ into $V$: $V = S_c \cup S_{md}$} \label{step:vertices}
            \State{Build edges $e$ by connecting vertices in $V$ if its corresponding cluster are neighbors (from $M_n$} \label{step:edges}
            \State{Estimate line set $S_l$ to fit the cluster set $S_c$ by $PCA$} \label{step:fitlines}
            \State{Determine intersection of feature lines \& their boundaries: $I_l = S_l \cap S_b$} \label{step:edgePoint}
            \For{i $\in I_l$} \label{step:checkandAdd}
                \For{j $\in V$}
                    \If{ distance from $I_l[i]$ to $V[j]$ larger than $d_{min}$}
                        \State{Add $I_l[i]$ into V}
                        \State{Edge $e$ = from $I_l[i]$ to its center vertex}
                        \State{Add $e$ into $E$}
                    \EndIf
                \EndFor
            \EndFor \label{step:endfor1}\\
            \Return $G = (V,E)$
        \EndProcedure
    \end{algorithmic}
\end{algorithm}

\textit{Algorithm \ref{alg:GraphEsti}} builds the graph from the cluster boundaries set $S_{bo}$ (from Algorithm \ref{alg:boundaryestimation}) and a distance threshold $d_{min}$. Firstly, the center points $S_c$ of clusters, neighbor matrix $M_n$, and border middle points $S_{md}$ are determined from step \ref{step:cenPoint2} to \ref{step:middlepoint}.
All the points in $S_c$ and $S_{md}$ are added into the vertices set $V$. The edge set is built by $M_n$ and $V$ in step \ref{step:edges}. After that, a set of feature's lines that are fit to the boundaries are calculated by \textit{PCA} method \cite{fukunaga2013introduction}. Step \ref{step:edgePoint} determines the intersection set $I_l$ of the feature's lines and their cluster boundary $S_b$.
From step \ref{step:checkandAdd} to \ref{step:endfor1}, there are dual loops to check whether the distance from a point in $I_l$ to a vertex in $V$ is longer than the threshold $d_{min}$. If yes, then that point and an edge, which connects it to its center point, are added into the vertex set $V$ and the edge set $E$, respectively. The algorithm outputs $G = (V,E)$, which is sent to \textit{Algorithm} \ref{alg:PathVisiAllEdge}.

\begin{algorithm}[ht]
\small
\caption{Variant Open CPP}\label{alg:PathVisiAllEdge}
    \begin{algorithmic}[1]
        \Procedure{Minimal Path}{$G = (\mathcal{V}, \mathcal{E}), v_s, v_t$}
            \State Initialize $\mathcal{G}_m, S_{mov}, S_{ov}$
            \State Find all odd vertices and add to $S_{ov}$.
            \If{$S_{ov} = \emptyset$} \label{step:eulerian}
                \State{Find shortest edge set $\mathcal{E}_e $ connecting $v_s$ and $v_t$
                \State Assign $\mathcal{G}_m = (\mathcal{V}, \mathcal{E_m} = \mathcal{E} \cup \mathcal{E}_e)$}, go to \ref{step:solEuPa}.
            \EndIf
            \If{($v_s, v_t$) $\subset$ $S_{ov}$} \label{step:bothinside}
                \State{$S_{mov} = S_{ov}\backslash \{v_s, v_t \}$, $G_m = G$, go to \ref{step:contoEulPath}.}
            \EndIf
            \If {$v_s \notin$ $S_{ov}$ and $v_t \in$ $S_{ov}$} \label{step:vtinside}
                \State{Select a vertex $v_c \in S_{ov}$, which creates the shortest path $\mathcal{E}_c$ connecting $v_s$ to it}
                \State {$S_{mov} = S_{ov}\backslash \{v_t, v_c \}$, $\mathcal{G}_m = (\mathcal{V}, \mathcal{E} \cup \mathcal{E}_c)$, go to \ref{step:contoEulPath}}
            \EndIf
            \If{$v_s \in$ $S_{ov}$ and $v_t \notin$ $S_{ov}$} \label{step:vsinside}
                \State{Select a vertex $v_c \in S_{ov}$, which creates the shortest path $\mathcal{E}_c = $ connecting $v_t$ to $v_c$}
                \State{$S_{mov} = S_{ov}\backslash \{v_s, v_c \}$, $\mathcal{G}_m = (\mathcal{V}, \mathcal{E} \cup \mathcal{E}_c)$, go to \ref{step:contoEulPath}}
            \EndIf
            \If{ $v_s \notin S_{ov}$ and $v_t \notin S_{ov}$} \label{step:allout}
                \State{Select two vertices $v_{c1}, v_{c2} \in S_{ov}$, which creates two paths $\mathcal{E}_{d1}, \mathcal{E}_{d2}$ that sum of them is shortest}
                \State{$S_{mov} = S_{ov}\backslash \{v_{c1}, v_{c2} \}$, $G_m = \{\mathcal{V}, \mathcal{E} \cup \mathcal{E}_{d1} \cup \mathcal{E}_{d2}\}$, go to \ref{step:contoEulPath}.}
            \EndIf
            \State \label{step:contoEulPath} Using $\mathcal{G}_m$, find a set of edges $\mathcal{E}_a$ whose sum is shortest to convert all vertices in $S_{mov}$ to even vertices.
            \State \label{step:solEuPa} Find the Eulerian path $P_{robot}$ from $\mathcal{G}_n = (\mathcal{V}, \mathcal{E}_m \cup \mathcal{E}_a)$ \\
            \Return $P_{robot}$
        \EndProcedure
    \end{algorithmic}
\end{algorithm}

After constructing the graph, the next step is to find the shortest path to inspect all the available steel bars. As the robot inspects the bridge parts step by step, the starting and ending point should be different. Moreover, the steel bridge structures are diverse, thus the input graph could consist of Euler circuit, Euler path, or none. Algorithm \ref{alg:PathVisiAllEdge} - Variant Open CPP (VOCPP) is proposed to solve the optimization problem that is based on \textit{Euler Theorem} \cite{edmonds1973matching} for Eulerian trail. The algorithm converts any input graph into an \textit{open CPP} graph by adding the shortest path into the input graph and makes a new one. The added paths are selected by using Dijkstra's algorithm \cite{cormen2009introduction}. The algorithm's output is the shortest path, which visits each edge at least once and starts and ends at the predefined positions.

The \textit{pseudo-code} is shown in Algorithm \ref{alg:PathVisiAllEdge}. The inputs are the graph $G=\{V,E\}$, starting and target vertex $v_s,v_t$. Step \ref{step:eulerian} checks whether an Eulerian circuit exists in the graph. If yes, a path generated by Dijkstra's algorithm with two vertices $v_s, v_t$ is added into the graph. After that, the algorithm will jump to step \ref{step:solEuPa}. If the odd vertex set $S_{ov}$ are not empty, step \ref{step:bothinside} checks whether $v_s, v_t$ belong to the set. If yes, both are popped out, and a set of shortest paths is added to convert all the remained odd nodes in $S_{ov}$ to even. Steps \ref{step:vtinside}, \ref{step:vsinside}, and \ref{step:allout} check whether the vertex $v_s$ or $v_t$ stays in the $S_{ov}$, or if both are out of the set. If any input vertex is odd, it will be popped out. Vertices in $S_{ov}$ that possess the shortest paths to the even input vertices are connected by a shortest path to convert the even input vertex into odd. After that, the selected vertex is also popped out of $S_{ov}$. All then go to step \ref{step:contoEulPath} to convert all the odd vertices in the remaining $S_{ov}$ into even vertices by Dijkstra's algorithm. Step \ref{step:solEuPa} will find the shortest path that traverses all edges of the graph by Fleury's algorithm \cite{thorup2000near}.

\subsection{Point Inside Boundary Check - PIBC}
After receiving the shortest path from the VOCPP algorithm, a motion planning is deployed to generate a path to control the robot to traverse all the defined edges. It plans the motion path for one edge each time. 

\begin{figure}[ht]
    \centering
    \setcounter{subfigure}{0}
    \subfigure[]{\includegraphics[height =0.2\linewidth]{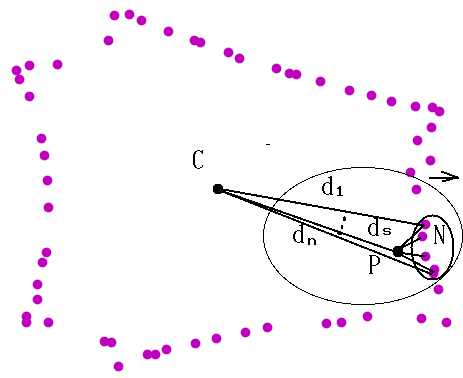}}
    \subfigure[]{\includegraphics[height =0.2\linewidth]{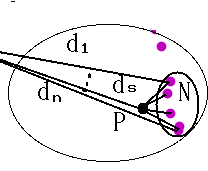}}
    \caption{Point inside the boundary check }
    \label{fig:Verify}
\end{figure}

In our scenario, the robot moves on a set of steel bars with assumption that there is no obstacle on it. The free configuration, however, is limited by the space of the steel bar surface. The serious damage can occur if the robot moves out of the steel bar edges. It is a hard constraint to construct a free configuration $C_{free}$ for the robot. To utilize all available configurations, the point cloud boundary is applied to build the obstacle-free configuration. 
After receiving the boundary set $S_{bo}$ from NCBE algorithm, a robot configuration $c_i$ belongs to the free configuration $C_{free}$ if all of its projected points $S_{pp}$ lie inside one of the steel bar boundaries $b_i \in S_{bo}$.

\begin{algorithm}[ht]
\small
\caption{Point Inside Boundary Check} \label{alg:veriSamp}
    \begin{algorithmic}[1]
        \Procedure{Sample Check}{$S_{bo}, c, P$}
            \State{Calculate the cluster center point set $c_p$ of each boundary in $S_{bo}$}
            \State{Project from configuration $c$ to the robot position set $PS$}
            \State{Initialize boolean set bs[len(PS)] = Fail}
            \For{$i \in (0,len(PS))$} \label{step:confstart}
                \State{Calculate distances $d_s$ from $PS[i]$ to $c_p$ }
                \State {Select $n$ center points, which are closest to $PS[i]$}
                \For{$p \in (0,n) $}
                    \State{Find $m_p$ closest points to $PS[i]$ in each $S_{bo}[p]$}
                    \For{$j \in m_p$}
                        \State{Calculate distance $d_{m_{pj}}$ from $m_{p}[j]$ to $c_p[p]$}
                        \If{$d_s < d_{n_{pj}}$}
                            \State{bs[i] = True}
                        \EndIf
                    \EndFor
                \EndFor
            \EndFor \label{step:confend}
            \For{$i \in (0,len(PS)$} \label{step:checkinside}
                \If{$bs[i] = Fail$}
                \Return{Fail}
                \EndIf
            \EndFor \\
            \Return {True}
        \EndProcedure
    \end{algorithmic}
\end{algorithm}

To determine whether a point lies inside a steel bar boundary, a geometric technique named \textit{center-closest points} is applied as shown in Fig. \ref{fig:Verify}. A sample point $P$ is considered inside the boundary if its distance $d_s$ to the center point $C$ is shorter than the distances $d_n$ of its closest neighbors to the center. The closest neighbor set $N$ is determined by finding the minimal Euclidean distance set.

The implementation of the technique is presented in \textit{Algorithm \ref{alg:veriSamp}}. The inputs are the boundary set $S_{bo}$, robot configurations $c$, and robot parameter set $P$. From step \ref{step:confstart} to \ref{step:confend}, the loop runs through all the robot configurations. For each configuration, it finds a set of clusters in which it possible belongs to. For each cluster, the technique \textit{center-closest points} is deployed to specify whether it lies inside that boundary. If the configuration stays inside only one of the cluster boundary, its corresponding boolean variable $bs[]$ will set \textit{True}.
Step \ref{step:checkinside} uses a \textit{for} loop to check whether there is any robot configuration not belonging to any cluster boundary. The algorithm returns \textit{True} if any configuration lies inside a boundary.

\section{Results}\label{sec_results}
In this section, the experiments and results are presented and discussed. The experiments were performed indoor with several samples of steel bridge structures. Due to the lack of equipment and limited indoor space, there are two sets of steel bridge structures were built to test the ARA robot on two transformations: \textit{Inchworm} and \textit{Mobile}, which are discussed in section \ref{sec:expe_setup}. The results are discussed in section \ref{sec:results}.

\subsection{Experiment Setup}\label{sec:expe_setup}
The experiments were implemented on the ARA robot version 1.0, which was derived from \cite{nguyen2020practical} with an additional camera module. An RGB-D camera (ASUS Xtion Pro Live) is attached to the robot for point cloud collection and visual inspection. The camera calibration's parameters were integrated from the method implemented by \cite{buiIRC2020Sort}. The robot was localized by \textit{aruco marker}, which was placed on the robot standing surface. We performed a geometric calculation to locate the position between the aruco marker and the robot base. Additionally, an Intel NUC 5 - Core i5 vPro was incorporated for employing the robotic operating system (ROS) as well as the inspection framework. 

The first experiment would test the control framework, which can switch the robot transformations, and control the robot to perform the inchworm jump, and run on smooth surfaces. Two steel slabs located perpendicularly from each other were set up for this task (Fig. \ref{fig:inch-wormjumpmotion}). The steel slabs are highly corroded to replicate steel defects. The second experiment was on the robot navigation as running on smooth surfaces. Several steel bridge structures such as \textit{K-}, \textit{L-}, \textit{I-}, \textit{T-}, and \textit{Cross-} shapes were assembled on the ground (Fig.\ref{fig:segment_boundary} a-e)  to simulate the typical steel bar structures. The following section describe the experiments and their results elaborately.

\subsection{Results}\label{sec:results}
The results of two experiments - control and navigation frameworks were presented and discussed. They are two critical portions, which constituted the autonomous system for the ARA robot.
\subsubsection{Switching control}
\begin{figure}[ht]
\centering
\includegraphics[height=0.35\linewidth]{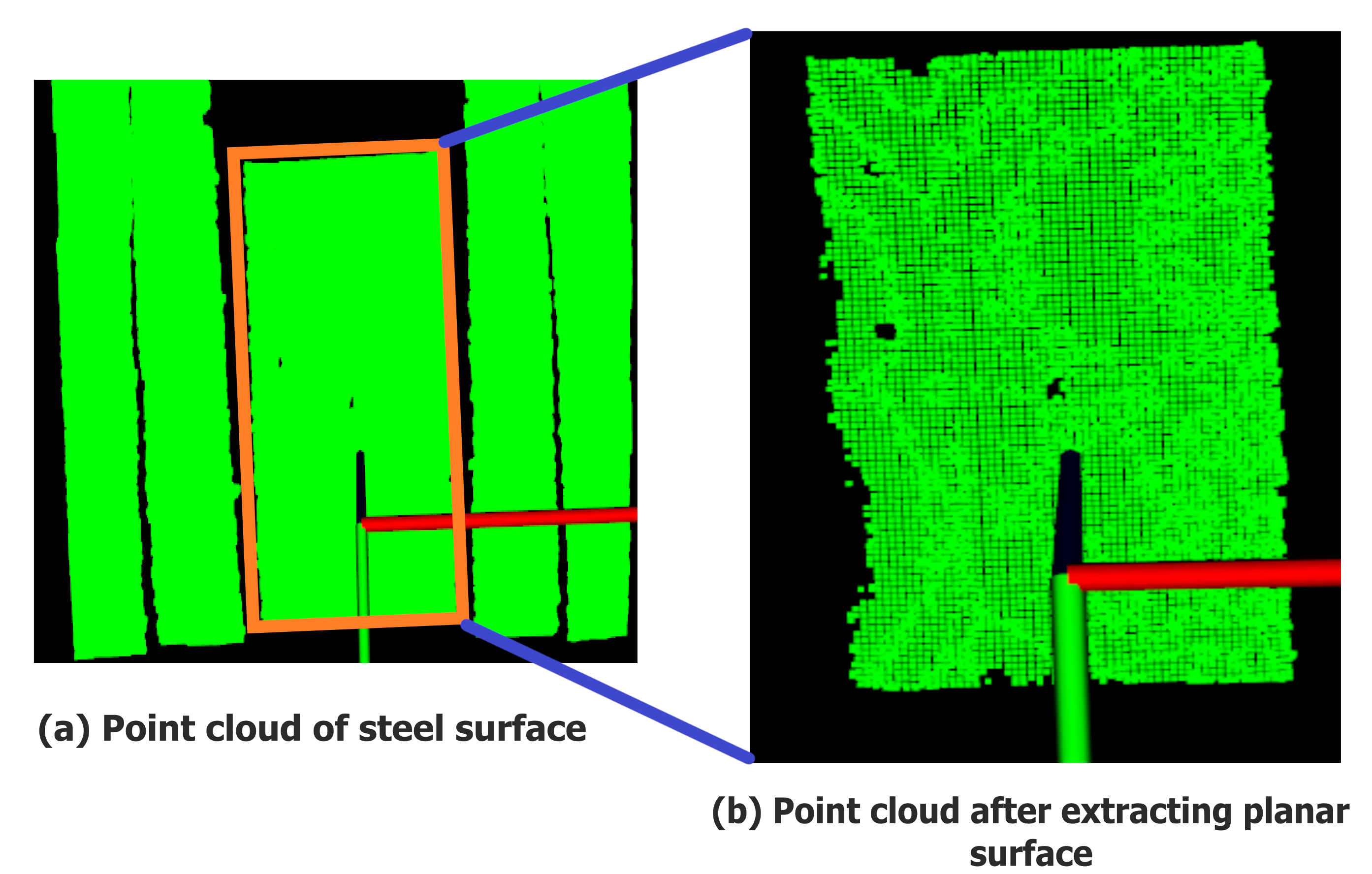}
\caption{Planar surface extraction from 3D point cloud of steel surface}
\label{fig:initialPointCloud}
\end{figure}

At starting, the PCL data of a steel bridge structure sample was collected from the robot camera. An example of the initial PCL was shown in Fig.\ref{fig:initialPointCloud}(a). After performing some pre-processing operations such as \textit{pass-through filtering} and \textit{downsampling}, the data was sent to \textit{plane detection} to extract the planar surface. The processed PCL was shown in Fig.\ref{fig:initialPointCloud}(b). The coordinate frame was also shown in the figure with \textit{x-}axis in red, \textit{y-}axis in green, and  \textit{z-}axis in blue.

\begin{figure}[ht]
\centering
\setcounter{subfigure}{0}
\subfigure[]{\includegraphics[width=0.2\linewidth, height =0.32\linewidth]{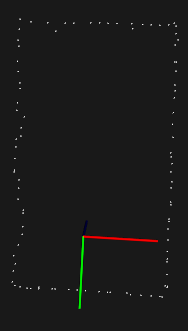}}
\subfigure[]{\includegraphics[width=0.22\linewidth, height =0.32\linewidth]{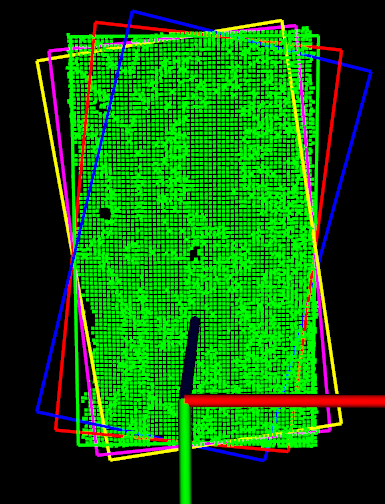}}
\subfigure[]{\includegraphics[width=0.20\linewidth, height =0.32\linewidth]{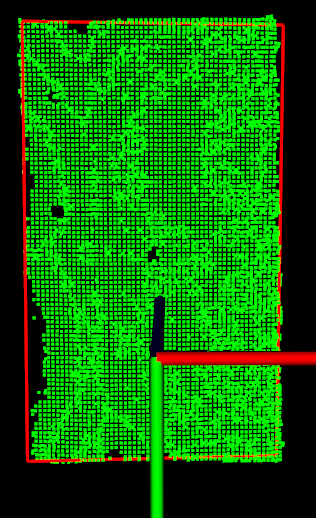}}
\subfigure[]{\includegraphics[width=0.24\linewidth, height =0.32\linewidth]{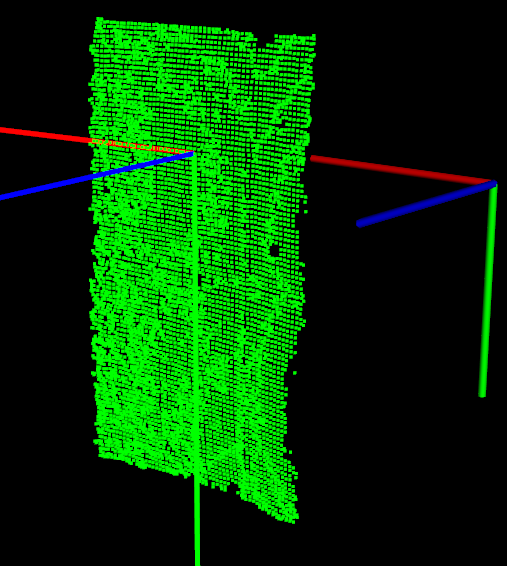}}
\caption{(a) Boundary set, (b) Area rectangle set, (c) The selected area rectangle, and (d) Pose estimation}
\label{fig:boundaryRectangle}
\end{figure}

After obtaining the planar surfaces,  the surface boundary points and \textit{Area availability} checking were performed by \textit{Algorithm \ref{alg:boundaryestimation}} and \textit{Algorithm \ref{alg:areaestimation}} on two different surfaces, one containing sufficient area for movement and the other without. Using \textit{Algorithm \ref{alg:boundaryestimation}}, two boundary points of the two different point cloud as shown in Fig.\ref{fig:boundaryRectangle}(a) and Fig.\ref{fig:boundaryRectangle}(c). 

The area availability check from \textit{Algorithm \ref{alg:areaestimation}} was employed using the boundary point estimated. The algorithm parameters were as following : $n = 5, m = 3 $ and $t = 0.02$. Five rectangles were estimated for the robot feet with this algorithm as shown in Fig. \ref{fig:boundaryRectangle}(b) in red, yellow, blue, green, and purple color. In Fig. \ref{fig:boundaryRectangle}(b), several corners of all the red, yellow, purple, and blue rectangles were outside of the point cloud area. It represented that these rectangles area were not sufficient enough for an inch-worm jump. Only the green rectangle was inside the point cloud, satisfying area requirement. The selected rectangle is shown in Fig.  \ref{fig:boundaryRectangle}(c) (in red color). Since there is enough area for an inch-worm jump, the variable $S_{am}$ is set to true by the algorithm. After that, the planar surface pose was estimated as shown in Fig. \ref{fig:boundaryRectangle}(d) with three orientations shown in red, green, and blue color, respectively on the point cloud surface.
 
\begin{figure}[ht]
\centering
\setcounter{subfigure}{0}
\subfigure[]{\includegraphics[width=0.35\linewidth, height = 0.2\linewidth]{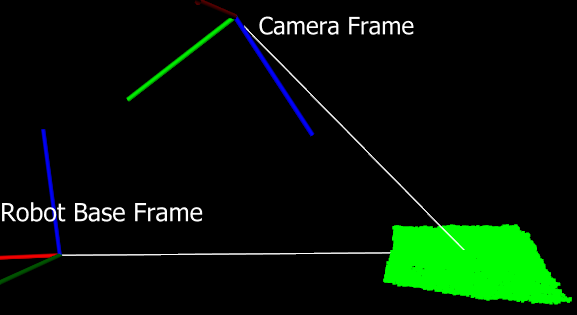}}
\subfigure[]{\includegraphics[width=0.33\linewidth, height = 0.2\linewidth]{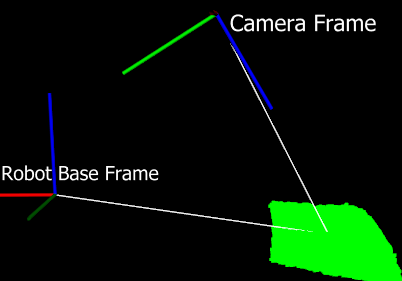}}
\caption{Surface Height Check (a) Same Height  \& (b) Different Height}
\label{fig:checkHeight}
\end{figure}

The surface pose was then transformed into the robot base frame. If the pose's height (corresponding to \textit{z-} axis in the robot base frame) was equal to robot base height, the value of $S_{hc} $ was set to \textit{True}, and the robot configured itself as \textit{mobile transformation}. Fig. \ref{fig:checkHeight}(b) represented another scenario as the point cloud is from a surface, which was $d=7cm$ lower than the robot base. In this case, the heights were different, then the returned value of variable $S_{hc}$ was \textit{false}, and robot performs \textit{inch-worm transformation} in the next step.

\subsubsection{Inchworm Transformation}
\textit{KDL} Inverse Kinematics and \textit{RRTConnect} motion planner in \textit{MoveIt} package were selected to implement the task, which calculated the robot inverse kinematics and generated a trajectory for an inch-worm jump from point $P_{conv}$ to the target plane. To do that, a primitive robot model was built in \textit{urdf} format, with the exact dimensions, joint types and limits to \textit{ARA} robot.  The generated trajectory - a ROS topic - was a set of robot joint angles, which the robot joints followed to reach the target pose. 

\begin{figure}[ht]
\centerline{\includegraphics[width=0.7\linewidth]{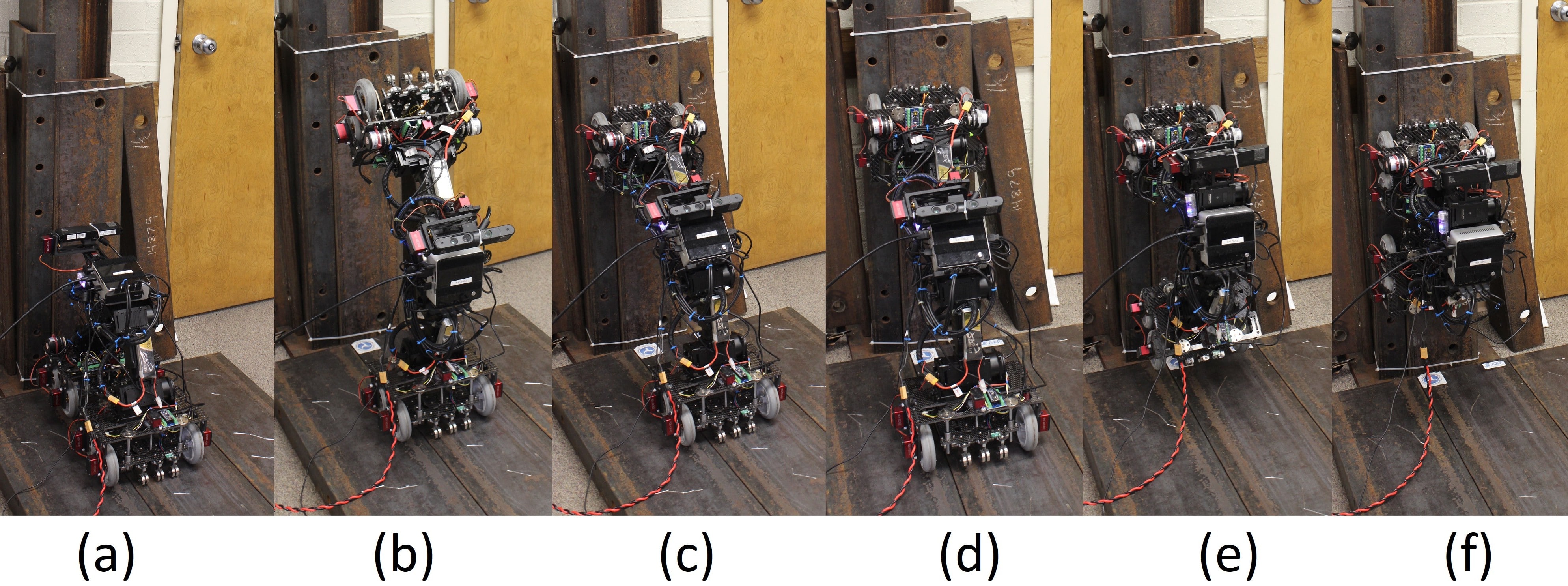}}
    \caption{\textit{inch-worm transformation}: a) magnetic array of second foot touched the base surface, b) first foot moved to convenient point, c) first foot reached target pose and touched the second surface, d) magnetic array of second foot was released, e) and f) second foot moved to target pose}
    \label{fig:inch-wormjumpmotion}
\end{figure}

The inch-worm performance of the robot was shown in Fig. \ref{fig:inch-wormjumpmotion}. In the beginning, the robot activated and lowed down the magnetic array to touch the steel surface; then it transformed from the mobile configuration to the convenient pose $P_{conv}$ by following a predefined trajectory as shown in Fig. \ref{fig:inch-wormjumpmotion}(a) and Fig. \ref{fig:inch-wormjumpmotion}(b). As reaching point $P_{conv}$, the robot started following the \textit{RRTConnect} trajectory. As the first foot reached the target surface, the robot switched both magnetic arrays working modes, the one on the first foot was changed to \textit{touched mode}, and the other was set to \textit{untouched mode} as shown in Fig. \ref{fig:inch-wormjumpmotion}(c)-(d). Next, the second robot foot transformed into the target plane as shown in Fig. \ref{fig:inch-wormjumpmotion}(e)-(f). The whole robot operation was filmed, and the video-clip was uploaded at \url{ https://youtu.be/SHk5IIOBRdA}, which was sped up three times than the experimental operation.  

\subsubsection{Navigation Framework}
\begin{figure*}[!]
    \centering
    \setcounter{subfigure}{0}
    \subfigure[]{\includegraphics[height =0.14\linewidth, width =0.17\linewidth, angle=90]{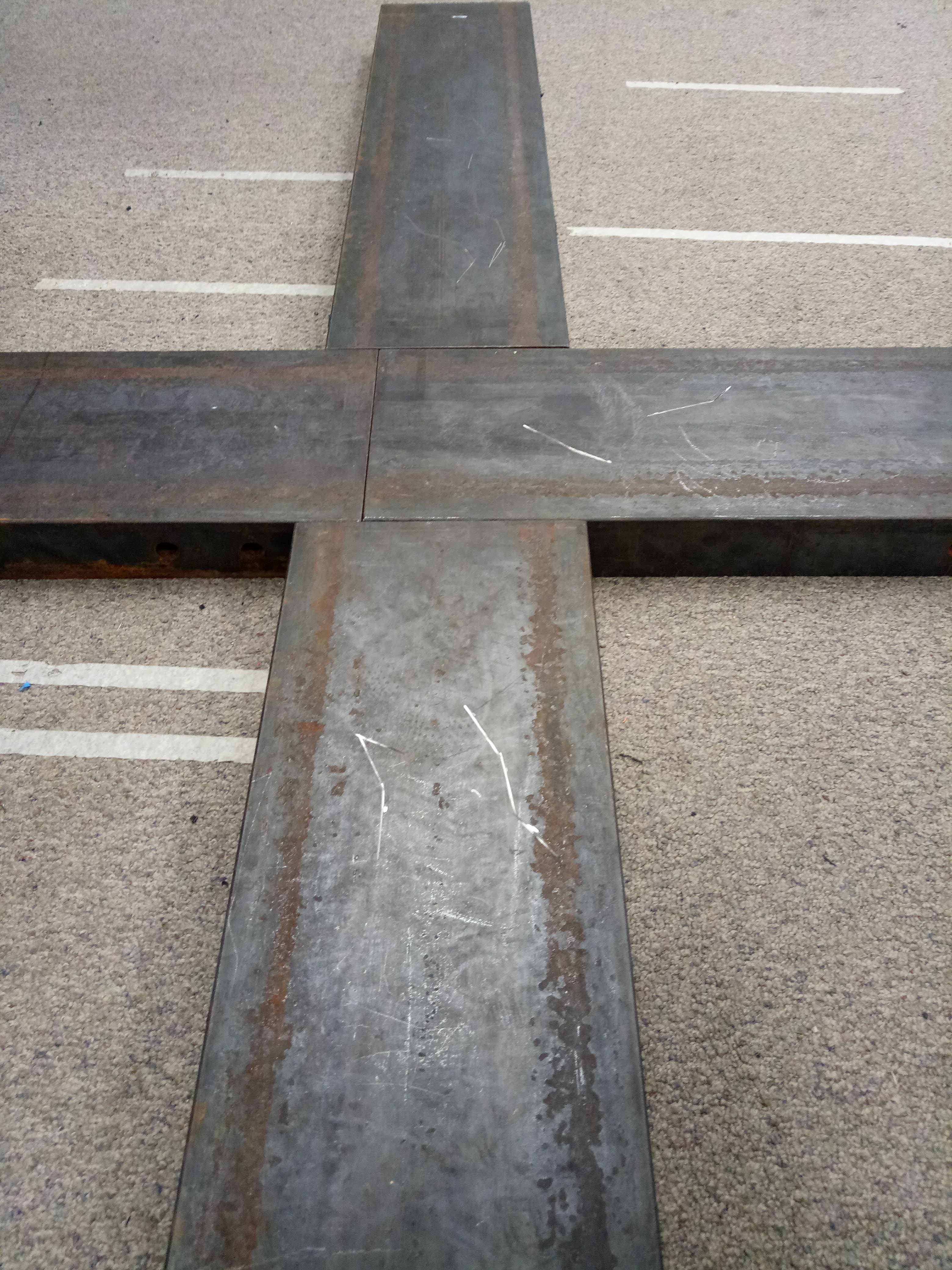}} \quad
    \subfigure[]{\includegraphics[height =0.14\linewidth, width =0.17\linewidth, angle=90] {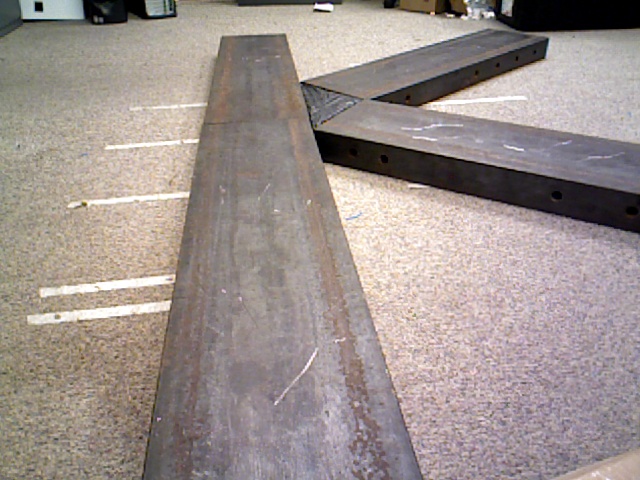}} \quad
    \subfigure[]{\includegraphics[height =0.14\linewidth, width =0.17\linewidth, angle=90] {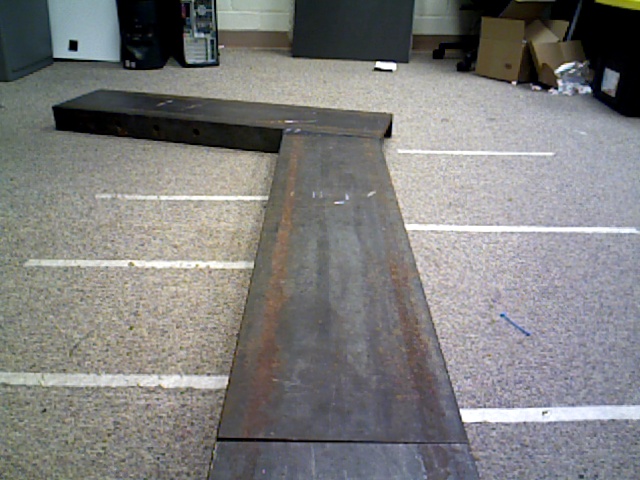}} \quad
    \subfigure[]{\includegraphics[height =0.14\linewidth, width =0.17\linewidth, angle=90] {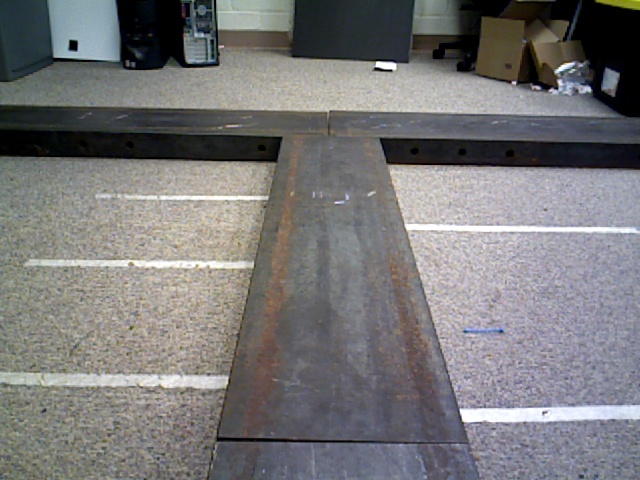}} \quad
    \subfigure[]{\includegraphics[height =0.14\linewidth, width =0.17\linewidth, angle=90] {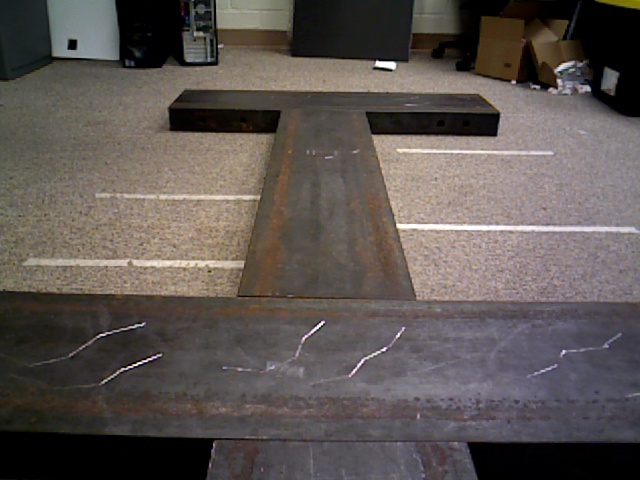}} \\
    \subfigure[]{\includegraphics[height =0.15\linewidth, width =0.17\linewidth]{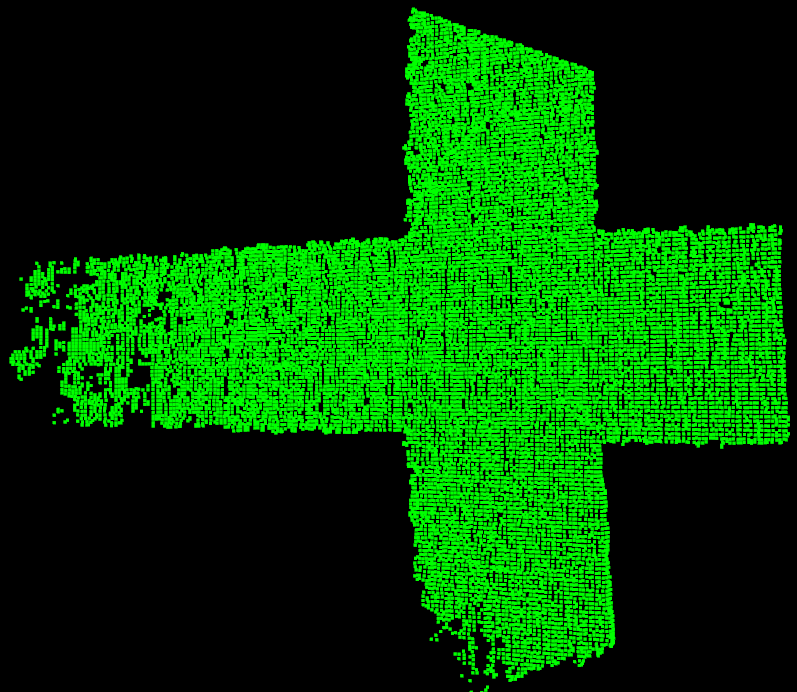}}
    \subfigure[]{\includegraphics[height =0.15\linewidth, width =0.17\linewidth]{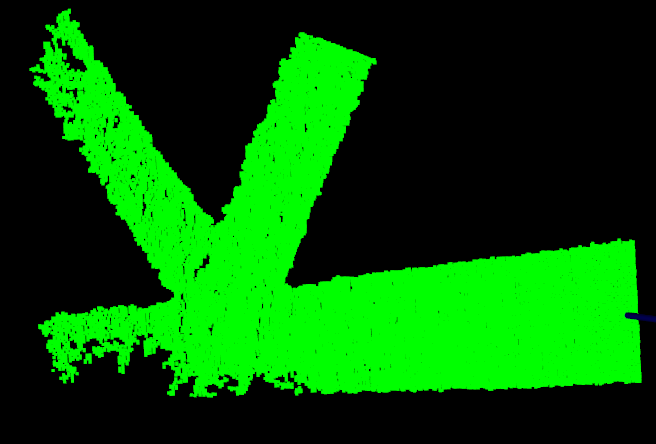}}
    \subfigure[]{\includegraphics[height =0.15\linewidth, width =0.17\linewidth]{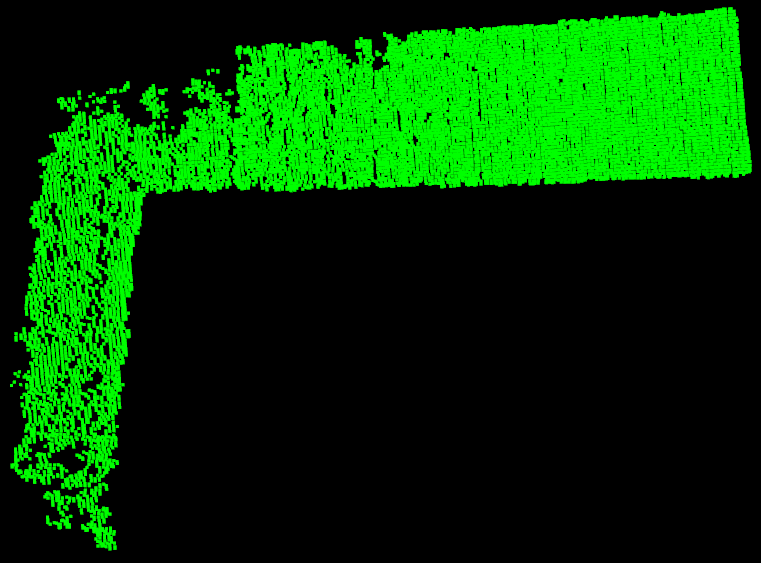}}
    \subfigure[]{\includegraphics[height =0.15\linewidth, width =0.17\linewidth]{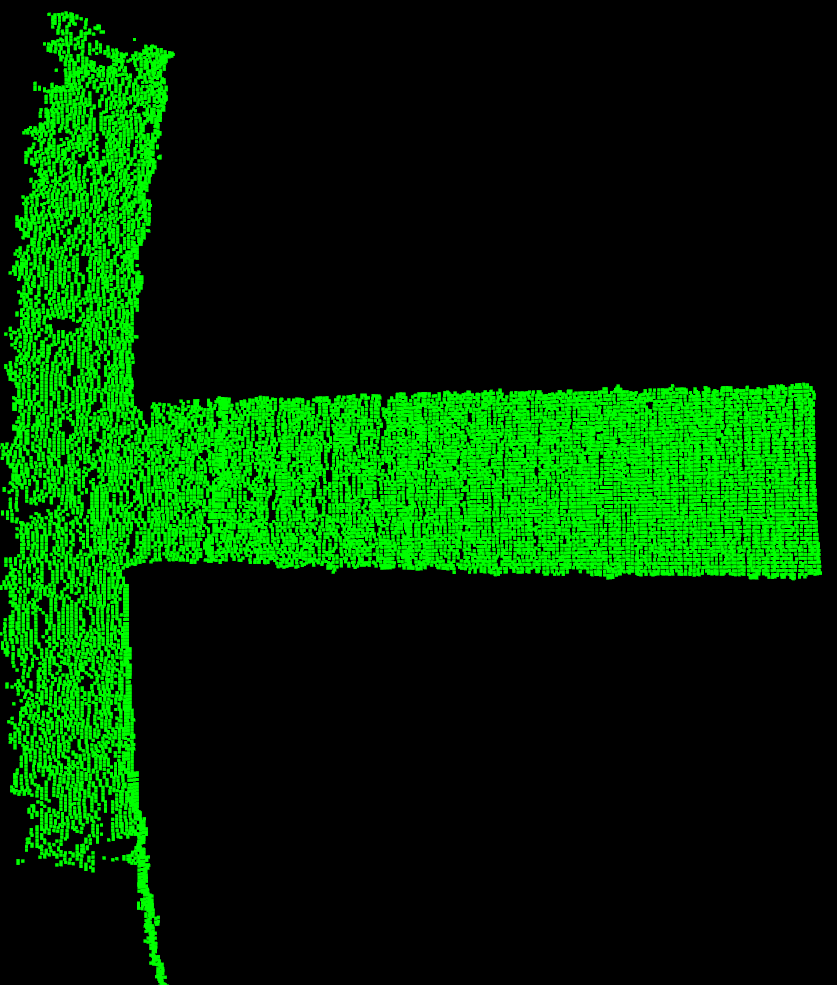}}
    \subfigure[]{\includegraphics[height =0.15\linewidth, width =0.17\linewidth]{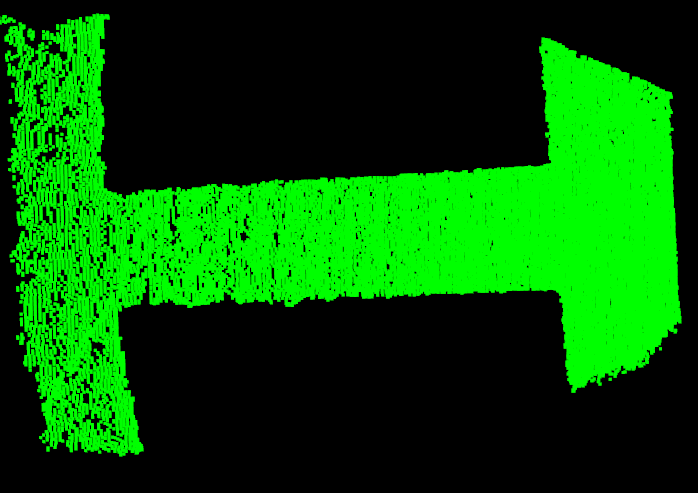}}\\
    \subfigure[]{\includegraphics[height =0.15\linewidth, width =0.17\linewidth]{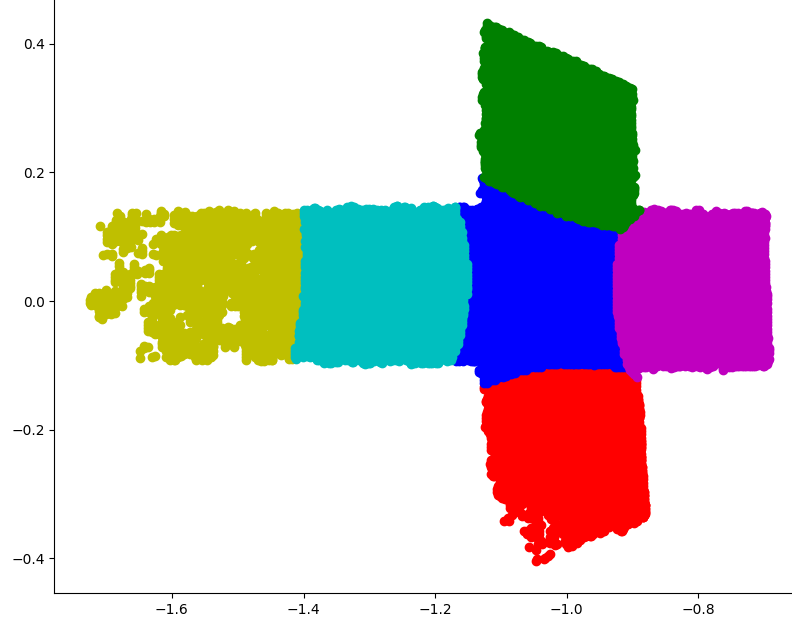}}
    \subfigure[]{\includegraphics[height =0.15\linewidth, width =0.17\linewidth]{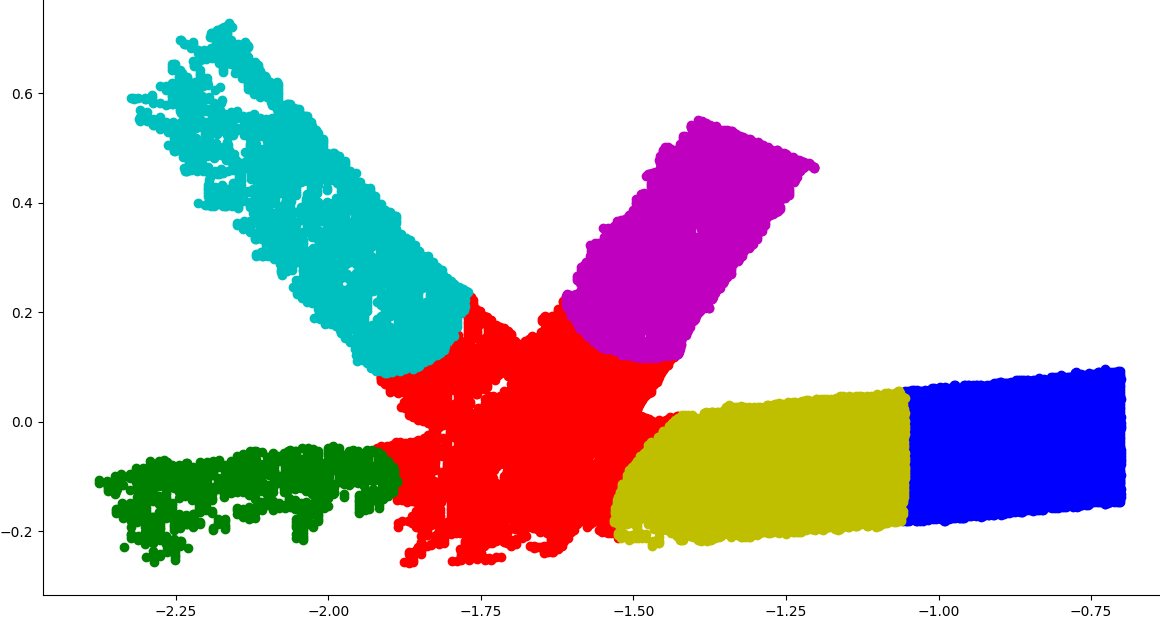}}
    \subfigure[]{\includegraphics[height =0.15\linewidth, width =0.17\linewidth]{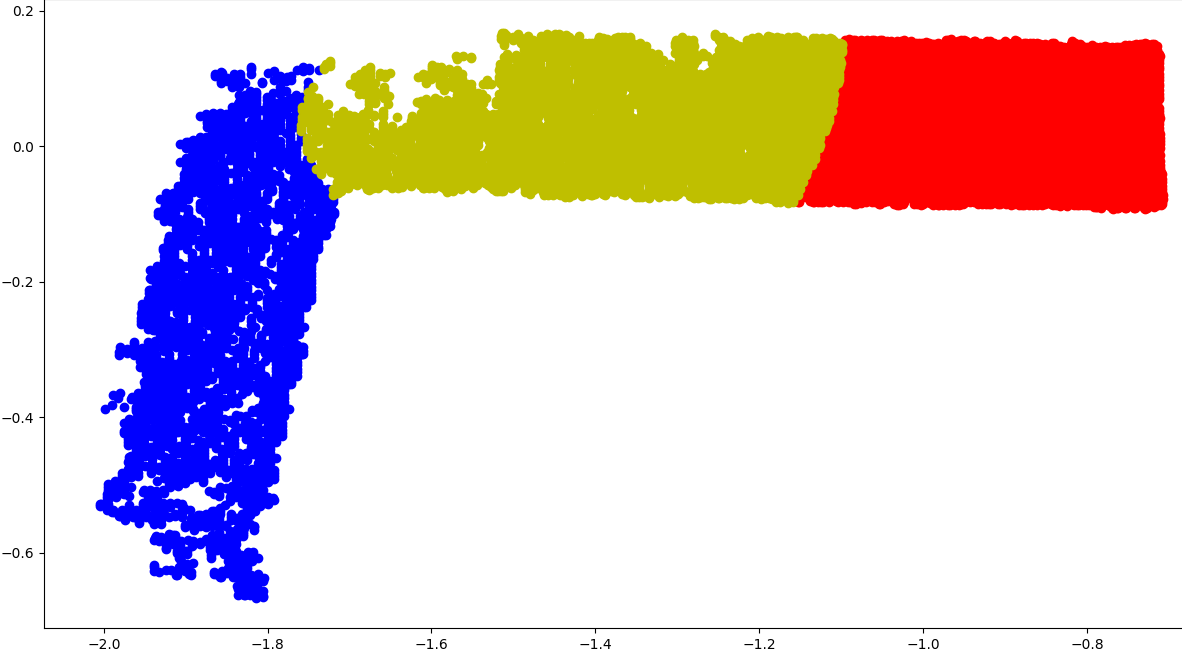}}
    \subfigure[]{\includegraphics[height =0.15\linewidth, width =0.17\linewidth]{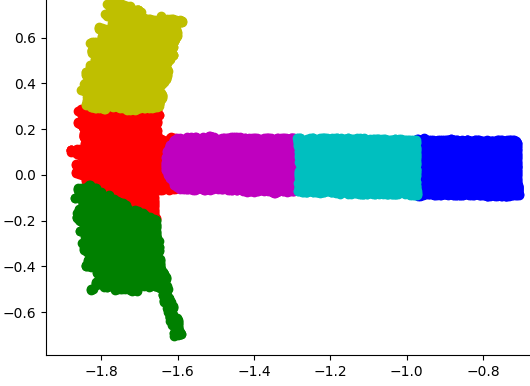}}
    \subfigure[]{\includegraphics[height =0.15\linewidth, width =0.17\linewidth]{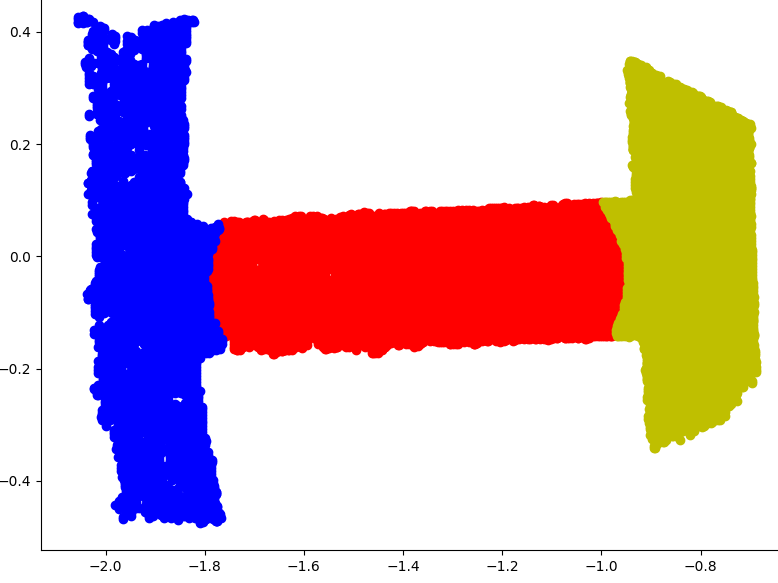}}\\
    \subfigure[]{\includegraphics[height =0.15\linewidth, width =0.17\linewidth]{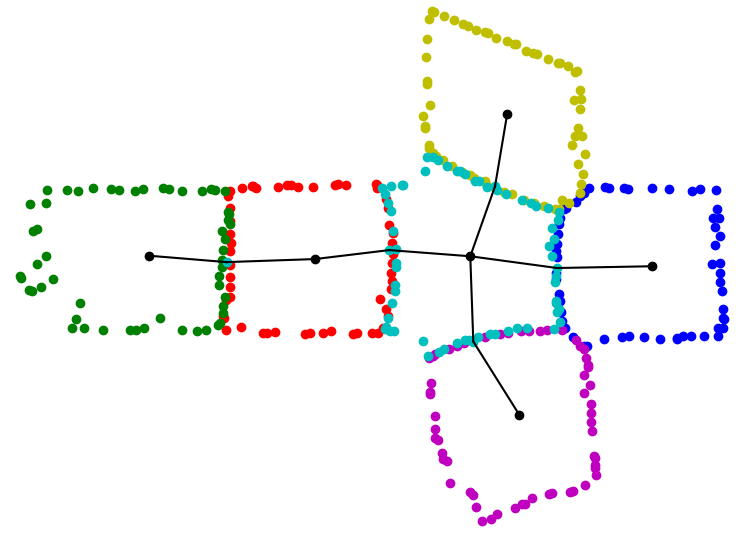}}
    \subfigure[]{\includegraphics[height =0.15\linewidth, width =0.17\linewidth]{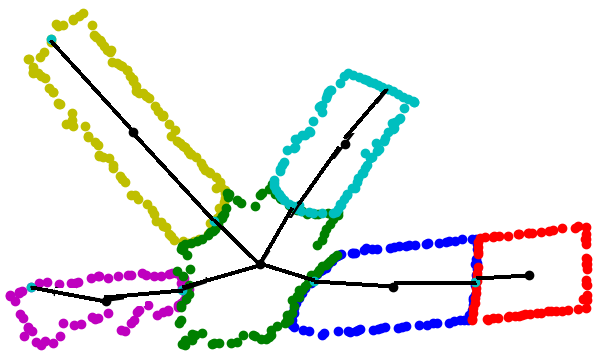}}
    \subfigure[]{\includegraphics[height =0.15\linewidth, width =0.17\linewidth]{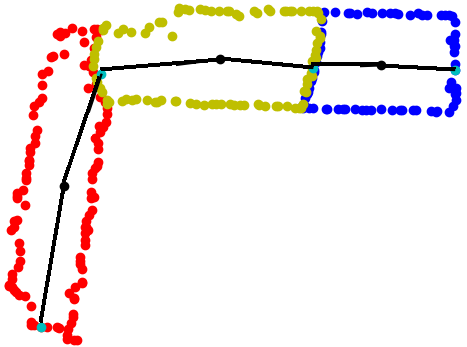}}
    \subfigure[]{\includegraphics[height =0.15\linewidth, width =0.17\linewidth]{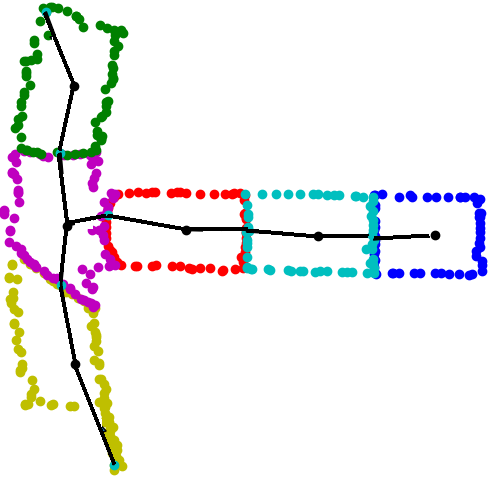}}
    \subfigure[]{\includegraphics[height =0.15\linewidth, width =0.17\linewidth]{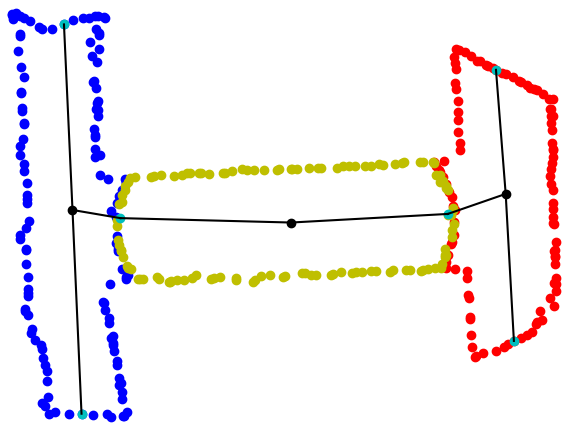}} \\
    \subfigure[]{\includegraphics[height =0.15\linewidth, width =0.17\linewidth]{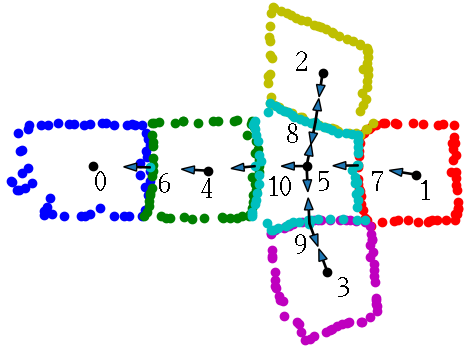}}
    \subfigure[]{\includegraphics[height =0.15\linewidth, width =0.17\linewidth]{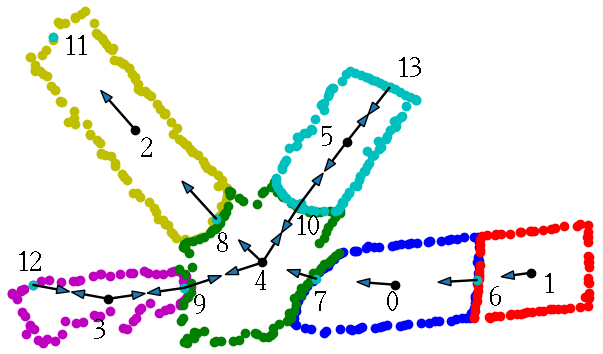}}
    \subfigure[]{\includegraphics[height =0.15\linewidth, width =0.17\linewidth]{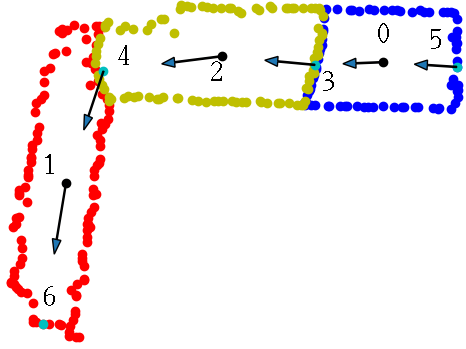}}
    \subfigure[]{\includegraphics[height =0.15\linewidth, width =0.17\linewidth]{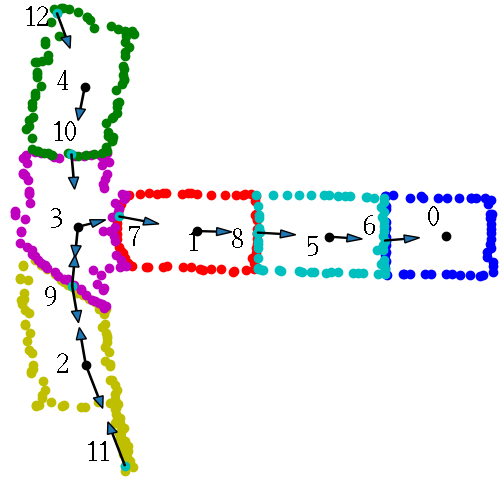}}
    \subfigure[]{\includegraphics[height =0.15\linewidth, width =0.17\linewidth]{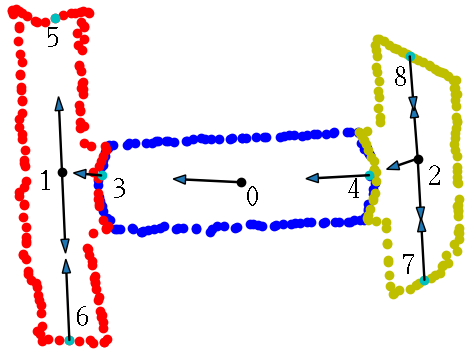}}
    \caption{The images of input structures (a-e), the corresponding point clouds (f-j), the segmentation (k-o), boundary estimation and graph construction (p-t), and the shortest path (u-y)}
    \label{fig:segment_boundary}
\end{figure*}

In this experiment, the four algorithms in section \ref{sec_navi} were tested on steel bridge structures such as \textit{L}-, \textit{Cross}-, \textit{T}-, \textit{K}, and \textit{I-} shapes. Due on the lab condition, we combined several steel bars on the ground to make the shape types mentioned as shown in the first row of Fig. \ref{fig:segment_boundary}. To present the experiment results succinctly, the structure images and PCL data were rotated 90 degree with the viewpoint from the right to left. 

\textbf{Structure Segmentation and Boundary Estimation}
\newline
The result of the segmentation algorithm was shown in Fig. \ref{fig:segment_boundary}. The images in the first and second rows were the RGB and point cloud data of \textit{Cross}-, \textit{K}-, \textit{L}-, \textit{T}-, and \textit{I}- shapes after filtering and projected into the robot coordinate frames. The view-point of camera was from right to left, and the farther to the camera, the more degrading of sensory data quality. The efficiency of \textit{Algorithm \ref{alg:PclSeg}} was shown in the third row of Fig. \ref{fig:segment_boundary}. The algorithm ran well and was able to segment properly the cross areas and the steel bar parts, except the \textit{I}- shape. The reason was that there were two cross areas in \textit{I}-shape structure, which made \textit{algorithm \ref{alg:PclSeg}} confuse. This problem will be improved in the future research. By a robust graph construction algorithm, however, the graph of \textit{I}- shape could be still built and helped generate the path for the robot. After segmenting, the clusters were sent to the  \textit{NCBE algorithm} \cite{bui2020control}, which worked efficiently to give back the boundaries for the cluster.

\textbf{Graph Construction and VOCPP}
\newline
Graph Construction (\textit{Algorithm \ref{alg:GraphEsti}}) and VOCPP (\textit{Algorithm \ref{alg:PathVisiAllEdge}}) processed the boundary data from the \textit{NCBE algorithm}, and outputted the result on the fourth and fifth rows, respectively. 
In the fourth row, the graph was built based on the center point, edge points of the boundary and the border points between the neighbor clusters. The graphs covered the steel bars' length for most structures, except the \textit{Cross}- structure (Fig. \ref{fig:segment_boundary}p). It was because of the distance from the center points to the corresponding edge points were too close, the depth sensor could cover all the distance. Therefore, it was no need an edge to connect the two points. In Fig. \ref{fig:segment_boundary}v, the edge (12-3) did not go along with the steel bar but crossing on one edge. It occurred because the data quality was not good, and influencing the line fitting algorithm. The same reason happened to the \textit{T}- structure in Fig. \ref{fig:segment_boundary}s.

The figures in the last row of Fig. \ref{fig:segment_boundary} show the shortest path generated by the \textit{VOCPP} algorithm. Due to the probabilistic feature of EM-GMM algorithm, the cluster indices changed each time running. Starting at a random vertex $v_s$, the robot followed the arrow lines to the next vertex, and comes back if there was a dead end. The route ended at the predefined ending vertex $v_t$, and the generated paths were optimal with shortest length. Again, due to the point cloud data quality, in Fig. \ref{fig:segment_boundary}v,x, the edges \textit{(3,12)}, \textit{(2,11)} were not possible for robot to traverse. To prevent the robot go on the edges, the motion planner was needed to handle this case.

\begin{figure}[ht]
    \centering
    \includegraphics[height =0.25\linewidth]{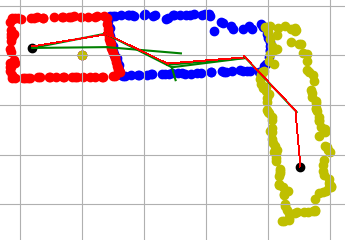}
    \caption{Result of Algorithm \ref{alg:veriSamp}}
    \label{fig:rrt}
\end{figure}

\textbf{Point Inside Boundary Check - PIBC and Path Planning}
\newline
Using PIBC algorithm with RRT motion planner, a motion path was generated as shown in Fig. \ref{fig:rrt} in \textit{L-}shape structure. The algorithm was significantly affected by the data quality, however, it still worked well to determine whether a robot configuration belong to the free space. To reduce the processing time for the robot, RRT  motion planning \cite{karaman2011sampling, sakai2018pythonrobotics} was deployed in the robot.

\section{Conclusion and Future Work}\label{sec_conclusion}
\subsection{Conclusion}
In this research, control and navigation frameworks were proposed for ARA robot. In control framework, a switching control mechanism for autonomous navigation of bridge-inspection robots was developed. The unique feature of switching in two modes enhanced the flexibility of navigation and inspection. The most significant part of this framework are two algorithms: \textit{Algorithm \ref{alg:boundaryestimation} - Non-convex Boundary Estimation (NCBE)} and \textit{Algorithm \ref{alg:areaestimation} - Area Availability} to detect and determine area, plane and height availability. 

The navigation framework was developed for ARA robot to run on mobile transformation. In this transformation, the robot needed to cross and inspect all the steel bars. The major portions of this framework were four algorithms, which could process the depth data, then outputted a traverse path for the robot. \textit{Algorithm \ref{alg:PclSeg} - Structure Segmentation} segmented the steel bar structures into two sets: steel bars and cross areas. Based on the segmentation result, the graph construction - \textit{Algorithm \ref{alg:GraphEsti}} built a graph that represents the bridge structure. The graph is inputted to \textit{Algorithm \ref{alg:PathVisiAllEdge} - VOCPP} to generate the shortest path for the robot to move and inspect all the available steel bars. \textit{Algorithm \ref{alg:veriSamp}} helped RRT path planning generate a trajectory, which the robot controller regulated the robot to follow.

The two proposed frameworks are crucial portions to build an autonomous framework, which makes ARA robot implementing the tasks on the real steel bridge structure. To get a full autonomous framework for ARA robot, there are still several problems, which needed to be solved and improved, as described in the following section.

\subsection{Future Work}
Due to the complexity of Inchworm manipulator-like structure, the motion planner \textit{RRTConnect} in \textit{Inchworm transformation} was not robust in calculating the inverse kinematics and generating the trajectory for the robot, and needed to redo sometimes. This limits the jumping operation of ARA robot in complex steel bridge structure. Further investigation of deployment in actual steel bridges, building a new motion planner for this robot, and optimization of inch-worm transformation is necessary as the next phase of the research. 

In the navigation framework, the efficiency of the algorithms needs to extend to process more bridge structures. The stability of the segmentation algorithm, which depends on the probability method, is not in well-operating and sometimes outputs inappropriate segmentation. To solve that, the formula \ref{Eq:optclus} could need to be refined, or a CNN needs to be developed to provide better results for this algorithm.

Moreover, the integration of multiple portions into a single framework to handle the real-world environment was the most challenging part of this research. It needs a cooperation of a team with interdisciplinary knowledge and skills to let ARA robot run on a real steel bridge.


\newpage
\small
\bibliography{ref}
\end{document}